\documentclass{article}
\usepackage[margin=1.5in]{geometry}
\usepackage[acronym,smallcaps]{glossaries}
\usepackage{enumitem}
\usepackage{graphicx}
\usepackage{longtable}
\usepackage{pifont}
\usepackage{listings}
\usepackage[T1]{fontenc}
\usepackage[latin1]{inputenc}
\usepackage[table]{xcolor}    

\usepackage{algorithm}
\usepackage{algpseudocode}
\usepackage{varwidth}

\title{A Hybrid Architecture for \\Multi-Party Conversational Systems}

\usepackage{authblk}
\author[1]{Maira Gatti de Bayser}
\author[1]{Paulo Cavalin}
\author[1]{Renan Souza}
\author[1]{Alan Braz}
\author[1]{Heloisa Candello}
\author[1]{Claudio Pinhanez}
\author[2]{Jean-Pierre Briot}
\affil[1]{IBM Research, Rio de Janeiro, Brazil}
\affil[2]{Sorbonne Universit'es, UPMC Univ Paris 06, CNRS, Laboratoire d'Informatique de Paris 6 (LIP6), Paris, France}

\newacronym{MPCS}{MPCS}{Multi-Party Conversational System}
\newacronym{NDS}{NDS}{Natural Dialogue Systems}
\makeglossaries

\begin{document}

\label{firstpage}
\maketitle

\begin{abstract}
Multi-party Conversational Systems are systems with natural language interaction between one or more people or systems. From the moment that an utterance is sent to a group, to the moment that it is replied in the group by a member, several activities must be done by the system: utterance understanding, information search, reasoning, among others. In this paper we present the challenges of designing and building multi-party conversational systems, the state of the art, our proposed hybrid architecture using both norms and machine learning and some insights after implementing and evaluating one on the finance domain. 
\end{abstract}

\section{Introduction}
\label{sec:intro}

Back to 42 BC, the philosopher Cicero has raised the issue that although there were many Oratory classes, there were none for Conversational skills \cite{Cicero}. He highlighted how important they were not only for politics, but also for educational purpose. Among other conversational norms, he claimed that people should be able to know when to talk in a conversation, what to talk depending on the subject of the conversation, and that they should not talk about themselves.

Norms such as these may become social conventions and are not learnt at home or at school.  Social conventions are dynamic and may change according to context, culture and language. In online communication, new commonsense practices are evolved faster and accepted as a norm \cite{Danescu-Niculescu-Mizil2013}, \cite{Jenks2014}.  There is not a discipline for that on elementary or high schools and there are few linguistics researchers doing research on this field.

On the other hand, within the Artificial Intelligence area, some Conversational Systems have been created in the past decades since the test proposed by Alan Turing in 1950. The test consists of a machine's ability to exhibit intelligent behavior equivalent to, or indistinguishable from that of a human \cite{Turing50}. Turing proposed that a human evaluator would judge natural language conversations between a human and a machine that is designed to generate human-like responses. Since then, many systems have been created to pass the Turing's test. Some of them have won prizes, some not \cite{Loebner1990}. Although in this paper we do not focus on creating a solution that is able to build conversational systems that pass the Turing's test, we focus on \gls{NDS}. From \cite{Berg2014}, \textit{"\gls{NDS} are systems that try to improve usability and user satisfaction by imitating human behavior"}. We refer to Conversational Systems as \gls{NDS}, where the dialogues are expressed as natural language texts, either from artificial intelligent agents (a.k.a. bots) or from humans.

That said, the current popular name to systems that have the ability to make a conversation with humans using natural language is \textit{Chatbot}. Chatbots are typically used in conversational systems for various practical purposes, including customer service or information acquisition. Chatbots are becoming more widely used by social media software vendors.  For example, Facebook\footnote{http://www.facebook.com} recently announced that it would make Facebook Messenger (its 900-million-user messaging app by 2016), into a full-fledged platform that allows businesses to communicate with users via chatbots.  Google is also building a new mobile-messaging service that uses artificial intelligence know-how and chatbot technology. In addition, according to the Wall Street Journal, there are more than 2 billion users of mobile apps. Still, people can be reluctant to install apps. So it is believed that social messaging can be a platform and chatbots may provide a new conversational interface for interacting with online services, as chatbots are easier to build and deploy than apps \cite{WSJ2016}.

China seems to be the place where chatbots adoption and use is most advanced today. For example, China's popular WeChat messaging platform can take payments, scan QR codes, and integrate chatbot systems. WeChat integrates e-mail, chat, videocalls and sharing of large multimedia files. Users can book flights or hotels using a mixed, multimedia interaction with active bots. WeChat was first released in 2011 by Tecent, a Chinese online-gaming and social-media firm, and today more than 700 million people use it, being one of the most popular messaging apps in the world (The Economist 2016). WeChat has a mixture of real-live customer service agents and automated replies  (Olson 2016).

Still, current existing chatbot engines do not properly handle a group chat with many users and many chatbots. This makes the chatbots considerably less social, which is a problem since there is a strong demand of having social chatbots that are able to provide different kinds of services, from traveling packages to finance advisors. This happens because there is a lack of methods and tools to design and engineer the coordination and mediation among chatbots and humans, as we present in Sections 2 and 3. In this paper, we refer to conversational systems that are able to interact with one or more people or chatbots in a multi-party chat as \textit{\gls{MPCS}}. Altogether, this paper is not meant to advance the state of the art on the norms for \gls{MPCS}. Instead, the main contributions of this paper are threefold:
\begin{enumerate}
\item We discuss the challenges of designing and building \gls{MPCS} (Section \ref{sec:challenges}),
\item We categorize the state of the art with works that tackle each of the challenges somehow (Section \ref{sec:conversational}), 
\item We present our hybrid conceptual architecture (Section \ref{sec:arch}), and insights and lessons learned after implementing and validating one on the finance domain (Section \ref{sec:eval}).
\end{enumerate}

We then present some discussion and future work in the last section.

\section{Challenges on Chattering}
\label{sec:challenges}

There are plenty of challenges in conversation contexts, and even bigger ones when people and machines participate in those contexts.  Conversation is a specialized form of interaction, which follows social conventions. Social interaction makes it possible to inform, context, create, ratify, refute, and ascribe, among other things, power, class, gender, ethnicity, and culture \cite{Jenks2014}. Social structures are the norms that emerge from the contact people have with others \cite{Parsons1937}, for example, the communicative norms of a negotiation, taking turns in a group, the cultural identity of a person, or power relationships in a work context. 

Conventions, norms and patterns from everyday real conversations are applied when designing those systems to result in adoption and match user's expectations. \cite{Ju:2008} describes implicit interactions in a framework of interactions between humans and machines. The framework is based on the theory of implicit interactions which posits that people rely on conventions of interaction to communicate queries, offers, responses, and feedback to one another. Conventions and patterns drive our expectations about interactive behaviors. This framework helps designers and developers create interactions that are more socially appropriate. According to the author, we have interfaces which are based on explicit interaction and implicit ones. The explicit are the interactions or interfaces where people rely on explicit input and output, whereas implicit interactions are the ones that occur without user awareness of the computer behavior. 

Social practices and actions are essential for a conversation to take place during the turn-by-turn moments of communication. \cite{Suchman2007} highlights that a distinguishing feature of ordinary conversation is "the local, moment-by-moment management of the distribution of turns, of their size, and what gets done in them, those things being accomplished in the course of each current speaker's turn." Management of turns and subject change in each course is a situation that occurs in real life conversations based on circumstances (internal and external) to speakers in a dialogue. Nowadays, machines are not prepared to fully understand context and change the course of conversations as humans. Managing dialogues with machines is challenging, which increases even more when more than one conversational agent is part of the same conversation. Some of those challenges in the dialogue flow were addressed by \cite{Zue2000}. According to them, we have system-initiative, user-initiative, and mixed-initiative systems.

In the first case, system-initiative systems restrict user options, asking direct questions, such as (Table \ref{tab:system}): "What is the initial amount of investment?" Doing so, those types of systems are more successful and easier to answer to. On the other hand, user-initiative systems are the ones where users have freedom to ask what they wish. In this context, users may feel uncertain of the capabilities of the system and starting asking questions or requesting information or services which might be quite far from the system domain and understanding capacity, leading to user frustration. There is also a mixed-initiative approach, that is, a goal-oriented dialogue which users and computers participate interactively using a conversational paradigm.  Challenges of this last classification are to understand interruptions, human utterances, and unclear sentences that were not always goal-oriented.

The dialog in Table \ref{tab:system} has the system initiative in a question and answer mode, while the one in Table \ref{tab:usersystem} is a natural dialogue system where both the user and the system take the initiative. If we add another user in the chat, then we face other challenges. 

\begin{table}[h]
  \caption{System initiative}
  \centering
    \begin{tabular}{|l|l|}
    \hline
\textbf{Turn} & \textbf{Utterance} \\\hline\hline
S & Which investment option do you choose? \\\hline
U & Dollar \\\hline
S & What is the initial amount of investment? \\\hline
U & R\$ 35,000 \\\hline
S & What is the time period of investment? \\\hline
U &  2 years \\\hline
S & The return of investment is R\$38,000 at the end. \\\hline
  \end{tabular}
   \vspace{-1\baselineskip}
  \label{tab:system}
\end{table}

\begin{table}[h]
  \caption{User initiative}
  \centering
    \begin{tabular}{|p{1cm}|p{10cm}|}
    \hline
 \textbf{Turn} & \textbf{Utterance} \\\hline\hline
U & What is an investment? \\\hline
S & To invest is to allocate money (or sometimes another resource, such as time) in the expectation of some benefit in the future. In finance, the expected future benefit from investment is a return. \\\hline
U & What are the types of investment? \\\hline
S & There are traditional and alternative investments. \\\hline
U & Which investments are traditional investments? \\\hline
S & Bonds, cash and real state. \\\hline
    \end{tabular}
   \vspace{-1\baselineskip}
  \label{tab:user}
\end{table}

\begin{table}[h]
\caption{Natural Dialogue with Mixed initiative}
\centering
\begin{tabular}{|p{1cm}|p{10cm}|}
 \hline
 \textbf{Turn} & \textbf{Utterance} \\\hline\hline
S & How can I help you? \\\hline
U & I would like to invest in Dollars, is it good? \\\hline
S & Oh, that would be a great choice. What would be the initial amount of investment and for how long would you keep the money invested? \\\hline
U & Maybe R\$35,000 in 2 years? \\\hline
S & OK, um, let me see? The return of investment is R\$38,000 at the end.\\\hline
\end{tabular}
\vspace{-1\baselineskip}
\label{tab:usersystem}
\end{table} 

In Table \ref{tab:multiparty}, line 4, the user U1 invites another person to the chat and the system does not reply to this utterance, nor to utterances on lines 6, 7 and 8 which are the ones when only the users (wife and husband) should reply to. On the other hand, when the couple agrees on the period and initial value of the investment (line 9), then the system S1 (at the time the only system in the chat) replies indicating that it will invite more systems (chatbots) that are experts on this kind of pair $<$ period, initial value $>$. They then join the chat and start interacting with each other. At the end, on line 17, the user U2 interacts with U1 and they agree with the certificate option. Then, the chatbot responsible for that, S3, is the only one that replies indicating how to invest.

Table \ref{tab:multiparty} is one example of interactions on which the chatbots require knowledge of when to reply given the context of the dialog. In general, we acknowledge that exist four dimensions of understanding and replying to an utterance in \gls{MPCS} which a chatbot that interacts in a multi-party chat group should fulfill: 

\begin{enumerate}
\item \textit{\textbf{What} is the message/utterance about?} This task means to recognize the utterance, such as the intent of the utterance, the entity making the utterance, and features of the entity, such as a time or initial value; 
\item \textit{\textbf{Who} should reply to the utterance?}  I.e., to whom it is addressed? Should it be to a user? Or should it be to a chatbot? 
\item \textit{\textbf{How} the reply should be built/generated?}  For example, the reply depends on an execution of an action to generate the reply, like computing a value or guiding a car. 
\item \textit{\textbf{When} should the reply be sent?} For instance, perhaps the reply needs to be sent within 2 minutes, 10 minutes, 1 day, or after someone/some chatbot in the chat, or before someone, some chatbot speaks, etc.
\end{enumerate}

In the next section we present the state of the art and how they fullfil some of these dimensions.

\begin{table}[h!]
\caption{MultiParty Conversation with Mixed initiative}
\centering
\begin{tabular}{|p{0.4cm}|p{1cm}|p{10cm}|}
\hline
&  \textbf{Turn} & \textbf{Utterance} \\\hline\hline
1 & S1 & How can I help you? \\\hline
2 & U1 & I would like to invest in Dollars, is it good? \\\hline
3 & S1 & Well, not so sure. What would be the initial amount of investment and for how long would you keep the money invested? \\\hline
4 & U1 & Actually, I am not sure. Let me invite my husband to this chat... \\\hline
5 & U2 & $<<$ U2 joins the chat $>>$ \\\hline
6 & U1 & Honey, for how long would you like to keep the money invested? \\\hline
7 & U2 & for 2 years \\\hline
8 & U1 & Right, so R\$ 35,000 for 2 years? \\\hline
9 & U2 & yes \\\hline
10 & S1 & Ok, in this case I will invite two experts to simulate for you. \\\hline
11 & S2 & $<<$ S2 joins the chat $>>$ \\\hline
12 & S3 & $<<$ S3 joins the chat $>>$ \\\hline
13 & S1 & Experts, can you simulate the return of investment for R\$ 35,000 in 2 years? \\\hline
14 & S2 & Sure, in the savings account the return at the end will be R\$ 38,000\\\hline
15 & S3 & Sure, in the certificate of deposit the return at the end will be R\$ 38,600 \\\hline
16 & S1 & Thanks. It looks like it is better to invest in the certificate of deposit. \\\hline
17 & U2 & Honey, lets go with the certificate. \\\hline
18 & U1 & Ok, it seems a good idea... \\\hline
19 & S3 & Sure, to start you can click here. \\\hline
\end{tabular}
\vspace{-1\baselineskip}
\label{tab:multiparty}
\end{table}

\section{Conversational Systems}
\label{sec:conversational}

In this section we discuss the state of the art on conversational systems in three perspectives: types of interactions, types of architecture, and types of context reasoning. Then we present a table that consolidates and compares all of them. 

ELIZA \cite{Weizenbaum1976} was one of the first softwares created to understand natural language processing. Joseph Weizenbaum created it at the MIT in 1966 and it is well known for acting like a psychotherapist and it had only to reflect back onto patient's statements. ELIZA was created to tackle five \textit{"fundamental technical problems"}: the identification of critical words, the discovery of a minimal context, the choice of appropriate transformations, the generation of appropriate responses to the transformation or in the absence of critical words, and the provision of an ending capacity for ELIZA scripts.

Right after ELIZA came PARRY, developed by Kenneth Colby, who is psychiatrist at Stanford University in the early 1970s. The program was written using the MLISP language (meta-lisp) on the WAITS operating system running on a DEC PDP-10 and the code is non-portable.  Parts of it were written in PDP-10 assembly code and others in MLISP. There may be other parts that require other language translators. PARRY was the first system to pass the Turing test  - the psychiatrists were able to make the correct identification only 48 percent of the time, which is the same as a random guessing.

A.L.I.C.E. (Artificial Linguistic Internet Computer Entity) \cite{Wallace2002} appeared in 1995 but current version utilizes AIML, an XML language designed for creating stimulus-response chat robots \cite{Wallace2006}.  A.L.I.C.E.  bot  has,  at  present,  more  than  40,000  categories  of  knowledge,  whereas  the  original  ELIZA  had  only  about  200.  The program is unable to pass the Turing test, as even the casual user will often expose its mechanistic aspects in short conversations.

Cleverbot (1997-2014) is a chatbot developed by the British AI scientist Rollo Carpenter. It passed the 2011 Turing Test at the Technique Techno-Management Festival held by the Indian Institute of Technology Guwahati. Volunteers participate in four-minute typed conversations with either Cleverbot or humans, with Cleverbot voted 59.3 per cent human, while the humans themselves were rated just 63.3 per cent human \cite{Aron2011}.

\begin{table}[h]%
\caption{Categories of Classical Chatbots per Interaction and Intentions}
\centering
\begin{tabular}{|p{1.7cm}|p{1cm}|p{1.5cm}|p{0.8cm}|p{1.5cm}|}
\hline
 & \multicolumn{2}{|p{3.2cm}|}{\textbf{Interactions}}  & \multicolumn{2}{|p{2.8cm}|}{\textbf{Intentions}}  \\ \hline  
 &    \textit{Dyadic}      &   \textit{Coordinated}    &  \textit{Goal  }   &  \textit{Non Goal}    \\ \hline
ELIZA &    \ding{108}       &      \ding{109}      &  \ding{119}     &     \ding{108}     \\ \hline
PARRY &       \ding{108}       &     \ding{109}    &  \ding{119}    &    \ding{108}      \\ \hline
A.L.I.C.E &      \ding{108}      &    \ding{119}     &  \ding{119}    &     \ding{108}    \\ \hline
Cleverbot &    \ding{108}        &   \ding{119}   &   \ding{109}   &      \ding{108}            \\ \hline
\end{tabular}
\vspace{-1\baselineskip}
\label{tab:categories1}
\end{table}%

\begin{table}[h]%
\caption{Categories of Classical Chatbots per Architectures and Context Reasoning}
\centering
\begin{tabular}{|p{1.7cm}|p{1cm}|p{0.8cm}|p{0.8cm}|p{0.8cm}|p{0.8cm}|p{0.8cm}|p{0.7cm}|p{0.5cm}|p{0.5cm}|p{0.7cm}|}
\hline
 &  \multicolumn{3}{|p{3.2cm}|}{\textbf{Architectures}} & \multicolumn{3}{|p{4.2cm}|}{\textbf{Context Reasoning}}  \\ \hline  
 &  \textit{Rule}     &  \textit{Data}     &   \textit{Hybrid} &  \textit{Rule}     &  \textit{Data}     &   \textit{Hybrid}       \\ \hline
ELIZA &      \ding{108}      &      \ding{109}       &   \ding{109} &   \ding{119}   &  \ding{109}  &  \ding{109}  \\ \hline
PARRY &    \ding{119}    &    \ding{108}         &     \ding{108}        &      \ding{109}       &     \ding{109} & \ding{109}    \\ \hline
A.L.I.C.E &       \ding{109}         &      \ding{108}      &      \ding{108} &  \ding{119}   &  \ding{109}  &  \ding{109}    \\ \hline
Cleverbot &      \ding{109}         &      \ding{108}      &   \ding{109}  & \ding{109} &   \ding{108}   &  \ding{109}   \\ \hline
\end{tabular}
\vspace{-1\baselineskip}
\label{tab:categories2}
\end{table}%

\subsection{Types of Interactions}

Although most part of the research literature focuses on the dialogue of two persons, the reality of everyday life interactions shows a substantial part of multi-user conversations, such as in meetings, classes, family dinners, chats in bars and restaurants, and in almost every collaborative or competitive environment such as hospitals, schools, offices, sports teams, etc. The ability of human beings to organize, manage, and (mostly) make productive such complex interactive structures which are multi-user conversations is nothing less than remarkable. The advent of social media platforms and messaging systems such as WhatsApp in the first 15 years of the 21st century expanded our ability as a society to have asynchronous conversations in text form, from family and friends chatgroups to whole nations conversing in a highly distributed form in social media \cite{Jenkins2013}.

In this context, many technological advances in the early 2010s in natural language processing (spearheaded by the IBM Watson's victory in Jeopardy \cite{Ferrucci2012}) spurred the availability in the early 2010s of text-based chatbots in websites and apps (notably in China \cite{Olson2016}) and spoken speech interfaces such as Siri by Apple, Cortana by Microsoft, Alexa by Amazon, and Allo by Google. However, the absolute majority of those chatbot deployments were in contexts of dyadic dialog, that is, a conversation between a single chatbot with a single user. Most of the first toolkits for chatbot design and development of this initial period implicit assume that an utterance from the user is followed by an utterance of the chatbot, which greatly simplifies the management of the conversation as discussed in more details later. Therefore, from the interaction point of view, there are two types: 1) one in which the chatbot was designed to chat with one person or chatbot, and 2) other in which the chatbot can interact with more than two members in the chat.  \\

\noindent \textit{Dyadic Chatbot} \\

A Dyadic Chatbot is a chatbot that does know \textbf{when} to talk. If it receives an utterance, it will always handle and try to reply to the received utterance. For this chatbot to behave properly, either there are only two members in the chat, and the chatbot is one of them, or there are more, but the chatbot replies only when its name or nickname is mentioned. This means that a dyadic chatbot does not know how to coordinate with many members in a chat group. It lacks the social ability of knowing when it is more suitable to answer or not. Also, note that we are not considering here the ones that would use this social ability as an advantage in the conversation, because if the chatbot is doing with this intention, it means that the chatbot was designed to be aware of the social issues regarding a chat with multiple members, which is not the case of a dyadic chatbot. Most existing chatbots, from the first system, ELIZA \cite{Weizenbaum1976}, until modern state-of-the-art ones fall into this category. \\

\noindent \textit{Multiparty Conversations} \\

In multiparty conversations between people and computer systems, natural language becomes the communication protocol exchanged not only by the human users, but also among the bots themselves. When every actor, computer or user, understands human language and is able to engage effectively in a conversation, a new, universal computer protocol of communication is feasible, and one for which people are extremely good at. 

There are many differences between dyadic and multiparty conversations, but chiefly among them is turn-taking, that is, how a participant determines when it is appropriate to make an utterance and how that is accomplished. There are many social settings, such as assemblies, debates, one-channel radio communications, and some formal meetings, where there are clear and explicit norms of who, when, and for long a participant can speak. 

The state of the art for the creation of chatbots that can participate on multiparty conversations currently is a combination of the research on the creation of chatbots and research on the coordination or governance of multi-agents systems. A definition that mixes both concepts herein present is: \textit{A \textbf{chatbot} is an \textbf{agent} that interacts through natural language}. Although these areas complement each other, there is a lack of solutions for creating multiparty-aware chatbots or governed chatbots, which can lead to higher degree of system trust. 

\begin{enumerate}[label=(\roman*), leftmargin=1cm]

\item  \textit{Multi-Dyadic Chatbots} \\

Turn-taking in generic, multiparty spoken conversation has been studied by, for example, Sacks et al. \cite{Sacks1974}. In broad terms, it was found that participants in general do not overlap their utterances and that the structure of the language and the norms of conversation create specific moments, called transition-relevance places, where turns can occur. In many cases, the last utterances make clear to the participants who should be the next speaker (selected-next-speaker), and he or she can take that moment to start to talk. Otherwise, any other participant can start speaking, with preference for the first starter to get the turn; or the current speaker can continue \cite{Sacks1974}. \\

A key part of the challenge is to determine whether the context of the conversation so far have or have not determined the next speaker. In its simplest form, a vocative such as the name of the next speaker is uttered. Also, there is a strong bias towards the speaker before the current being the most likely candidate to be the next speaker. \\

In general the detection of transition-relevance places and of the selected-next-speaker is still a challenge for speech-based machine conversational systems. However, in the case of text message chats, transition-relevance places are often determined by the acting of posting a message, so the main problem facing multiparty-enabled textual chatbots is in fact determining whether there is and who is the selected-next-speaker. In other words, \textit{chatbots have to know when to shut up}. Bohus and Horowitz \cite{Bohus2011} have proposed a computational probabilistic model for speech-based systems, but we are not aware of any work dealing with modeling turn-taking in textual chats. \\

\item  \textit{Coordination of Multi-Agent Systems} \\

A multi-agent system (MAS) can be defined as a computational environment in which individual software agents interact with each other, in a cooperative manner, or in a competitive manner, and sometimes autonomously pursuing their individual goals. During this process, they access the environment's resources and services and occasionally produce results for the entities that initiated these software agents. As the agents interact in a concurrent, asynchronous and decentralized manner, this kind of system can be categorized as a complex system \cite{Jennings2001}. \\

Research in the coordination of multi-agent systems area does not address coordination using natural dialogue, as usually all messages are structured and formalized so the agents can reason and coordinate themselves.  On the other hand, chatbots coordination have some relations with general coordination mechanisms of multi-agent systems in that they specify and control interactions between agents. However, chatbots coordination mechanisms is meant to regulate interactions and actions from a social perspective, whereas general coordination languages and mechanisms focus on means for expressing synchronization and coordination of activities and exchange of information, at a lower computational level.\\

In open multi-agent systems the development takes place without a centralized control, thus it is necessary to ensure the reliability of these systems in a way that all the interactions between agents will occur according to the specification and that these agents will obey the specified scenario. For this, these applications must be built upon a law-governed architecture.  \\

Minsky published the first ideas about laws in 1987 \cite{Minsky1987}. Considering that a law is a set of norms that govern the interaction, afterwards, he published a seminal paper with the Law-Governed Interaction (LGI) conceptual model about the role of interaction laws on distributed systems \cite{Minsky2000}. Since then, he conducted further work and experimentation based on those ideas \cite{Minsky2003}.  Although at the low level a multiparty conversation system is a distributed system and the LGI conceptual model can be used in a variety of application domains, it is composed of abstractions basically related to low level information about communication issues of distributed systems (like the primitives \textit{disconnected}, \textit{reconnected}, \textit{forward}, and \textit{sending} or \textit{receiving of messages}), lacking the ability to express high level information of social systems. \\

Following the same approach, the Electronic Institution (EI) \cite{Esteva2003} solution also provides support for interaction norms. An EI has a set of high-level abstractions that allow for the specification of laws using concepts such as agent roles, norms and scenes. \\

Still at the agent level but more at the social level, the XMLaw description language and the M-Law framework \cite{Paes2005} \cite{Paes2007} were proposed and developed to support law-governed mechanism. They implement a law enforcement approach as an object-oriented framework and it allows normative behavior through the combination between norms and clocks. The M-Law framework \cite{Paes2007} works by intercepting messages exchanged between agents, verifying the compliance of the messages with the laws and subsequently redirecting the message to the real addressee, if the laws allow it. If the message is not compliant, then the mediator blocks the message and applies the consequences specified in the law, if any. They are called laws in the sense that they \textbf{enforce} the norms, which represent what can be done (permissions), what cannot be done (prohibitions) and what must be done (obligations). \\

\item \textit{Coordinated Aware Chatbots in a Multiparty Conversation} \\

With regard to chatbot engines, there is a lack of research directed to building coordination laws integrated with natural language. To the best of our knowledge, the architecture proposed in this paper is the first one in the state of the art designed to support the design and development of coordinated aware chatbots in a multiparty conversation. \\

\end{enumerate} 

\subsection{Types of Architectures}

There are mainly three types of architectures when building conversational systems: totally rule-oriented, totally data-oriented, and a mix of rules and data-oriented. \\

\noindent \textit{Rule-oriented} \\

A rule-oriented architecture provides a manually coded reply for each recognized utterance.  Classical examples of rule-based chatbots include Eliza and Parry. Eliza could also extract some words from sentences and then create another sentence with these words based on their syntatic functions. It was a rule-based solution with no reasoning. Eliza could not "understand" what she was parsing. More sophisticated rule-oriented architectures contain grammars and mappings for converting sentences to appropriate sentences using some sort of knowledge. They can be implemented with propositional logic or first-order logic (FOL). Propositional logic assumes the world contains facts (which refer to events, phenomena, symptoms or activities). Usually, a set of facts (statements) is not sufficient to describe a domain in a complete manner. On the other hand, FOL assumes the world contains Objects (e.g., people, houses, numbers, etc.), Relations (e.g. red, prime, brother of, part of, comes between, etc.), and Functions (e.g. father of, best friend, etc.), not only facts as in propositional logic. Moreover, FOL contains predicates, quantifiers and variables, which range over individuals (which are domain of discourse).

Prolog (from French: \textit{Programmation en Logique}) was one of the first logic programming languages (created in the 1970s), and it is one of the most important languages for expressing phrases, rules and facts. A Prolog program consists of logical formulas and running a program means proving a theorem. Knowledge bases, which include rules in addition to facts, are the basis for most rule-oriented chatbots created so far.

In general, a rule is presented as follows:
\begin{equation}
if <premise> then <conclusion>
\end{equation}

Prolog made it possible to perform the language of Horn clauses (implications with only one conclusion). The concept of Prolog is based on predicate logic, and proving theorems involves a resolute system of denials. Prolog can be distinguished from classic programming languages due to its possibility of interpreting the code in both a procedural and declarative way. Although Prolog is a set of specifications in FOL, it adopts the closed-world assumption, i.e. all knowledge of the world is present in the database. If a term is not in the database, Prolog assumes it is false. 

In case of Prolog, the FOL-based set of specifications (formulas) together with the facts compose the knowledge base to be used by a rule-oriented chatbot. However an Ontology could be used. For instance, OntBot \cite{AlZubaide2011} uses  mapping technique to transform ontologies and knowledge into relational database and then use that knowledge to drive its chats. One of the main issues currently facing such a huge amount of  ontologies  stored  in  a  database  is  the  lack  of  easy  to  use  interfaces  for  data  retrieval,  due  to  the  need  to  use  special  query  languages  or  applications.  

In rule-oriented chatbots, the   degree   of   intelligent   behavior   depends  on  the  knowledge  base size and quality (which represents the  information  that  the  
chatbot  knows),  poor  ones  lead  to  weak  chatbot  responses  while  good ones do the opposite. However, good knowledge bases may require years to be created, depending on the domain. \\

\noindent \textit{Data-oriented} \\
As opposed to rule-oriented architectures, where rules have to be explicitly defined, data-oriented architectures are
based on learning models from samples of dialogues, in order to reproduce the behavior of the interaction that are
observed in the data. Such kind of learning can be done by means of machine learning approach, or
by simply extracting rules from data instead of manually coding them. 

Among the different technologies on which these system can be based, we can highlight classical information 
retrieval algorithms, neural networks \cite{Bordes2016}, Hidden Markov Models (HMM) \cite{Wright1998},
and Partially Observable Markov Decision Process (POMDP) \cite{Roy2000}. 
Examples include Cleverbot and Tay \cite{Lee2016}. Tay was a chatbot developed by Microsoft that after one day 
live learning from interaction with teenagers on Twitter, started replying impolite utterances. Microsoft has 
developed others similar chatbots in China (Xiaoice\footnote{http://www.msxiaoice.com/}) and in Japan 
(Rinna\footnote{http://rinna.jp/}). Microsoft has not associated its publications with these chatbots, 
but they have published a data-oriented approach\cite{Liu2015} that proposes a unified multi-turn multi-task 
spoken language understanding (SLU) solution capable of handling multiple context sensitive classification 
(intent determination) and sequence labeling (slot filling) tasks simultaneously. The proposed architecture is 
based on recurrent convolutional neural networks (RCNN) with shared feature layers and globally normalized sequence 
modeling components.

A survey of public available corpora for can be found in \cite{Serban2017}. A corpus can be classified into
different categories, according to: the type of data, whether it is spoken dialogues, transcripts of
spoken dialogues, or directly written; the type of interaction, if it is human-human or human-machine;
and the domain, whether it is restricted or unconstrained. Two well-known corpora are the Switchboard
dataset, which consists of transcripts of spoken, unconstrained, dialogues, and the set of tasks for the
Dialog State Tracking Challenge (DSTC), which contain more constrained tasks, for instance the restaurant
and travel information sets.\

\noindent \textit{Rule and Data-oriented} \\

The  model  of  learning  in  current A.L.I.C.E. \cite{Wallace2006}  is  incremental or/and interactive  learning  because  a  person monitors  the  robot's conversations and creates new AIML content to make the responses more appropriate, accurate, believable, "human", or whatever he/she intends. There are algorithms  for  automatic  detection  of  patterns  in  the  dialogue  data and this  process provides  the  person  with  new  input  patterns  that  do  not  have specific replies yet, permitting a process of almost continuous supervised refinement of the bot.
 
As already mentioned, A.L.I.C.E. consists of roughly 41,000 elements called categories which is the basic unit of knowledge in AIML. Each category consists of  an  input  question,  an  output  answer,  and  an  optional  context.  The  question,  or  stimulus, is called the pattern. The answer, or response, is called the template. The two types of optional context are called \textit{that} and \textit{topic}. The keyword \textit{that} refers to the robot's previous utterance. The AIML pattern language consists only of words, spaces, and the wildcard symbols "\_" and "*". The words may consist only of letters and numerals. The pattern language is case invariant. Words are separated by a single space, and the wildcard characters function like words, similar to the initial pattern matching strategy of the Eliza system. More  generally,  AIML  tags  transform  the  reply into a mini computer program which can save data, activate other programs, give  conditional  responses,  and  recursively  call  the  pattern  matcher  to  insert  the  responses  from  other  categories.  Most  AIML  tags  in  fact  belong  to  this  template  side sublanguage \cite{Wallace2006}.

AIML language allows:

\begin{enumerate}
\item  Symbolic reduction: Reduce complex grammatical forms to simpler ones.
\item  Divide and conquer: Split an input into two or more subparts, and combine the 
responses to each.
\item  Synonyms: Map different ways of saying the same thing to the same reply.
\item  Spelling or grammar corrections: the bot both corrects the client input and acts as a language tutor.
\item  Detecting keywords anywhere in the input that act like triggers for a reply.
\item  Conditionals: Certain forms of branching to produce a reply.
\item  Any combination of (1)-(6).
\end{enumerate}

When the bot chats with multiple clients, the predicates are stored relative to each client ID. For example, the markup \textit{$<$set name$=$"name"$>$Matthew$<$/set
$>$} stores the string \textit{Matthew}  under  the  predicate  named  \textit{"name"}.  Subsequent  activations  of  \textit{$<$get name="name"$>$} return \textit{"Matthew"}.
In addition, one  of  the  simple  tricks  that  makes  ELIZA and A.L.I.C.E.  so  believable  is  a  pronoun  swapping  substitution. For instance: \\

\noindent U: My husband would like to invest with me.\\
\noindent S: Who else in your family would like to invest with you? 

\subsection{Types of Intentions}
According to the types of intentions, conversational systems can be classified into two 
categories: a) goal-driven or task oriented, and b) non-goal-driven or end-to-end systems.

In a goal-driven system, the main objective is to interact with the user so that back-end tasks,
which are application specific, are executed by a supporting system.
As an example of application we can cite technical support systems, for instance air ticket
booking systems, where the conversation system must interact with the user until all the 
required information is known, such as origin, destination, departure date and return date, 
and the supporting system must book the ticket. The most widely used approaches for developing
these systems are Partially-observed Decision Processes (POMDP) \cite{Roy2000}, Hidden Markov Models 
(HMM) \cite{Wright1998}, and more recently, Memory Networks \cite{Bordes2016}. Given that
these approaches are data-oriented, a major issue is to collect a large corpora of annotated 
task-specific dialogs. For this reason, it is not trivial to transfer the knowledge from 
one to domain to another. In addition, it might be difficult to scale up to larger sets of
tasks.

Non-goal-driven systems (also sometimes called reactive systems), on the other hand, generate utterances in accordance to user input,
e.g. language learning tools or computer games characters. These systems have become
more popular in recent years, mainly owning to the increase of popularity of Neural 
Networks, which is also a data-oriented approach. The most recent state of the art to develop such systems 
have employed Recurrent Neural Networs (RNN) 
\cite{Serban2015}, Dynamic Context-Sensitive Generation \cite{Sordoni2015}, and 
Memory Networks \cite{Sukhbaatar2015}, just to name a few. Nevertheless, probabilistic methods
such as Hidden Topic Markov Models (HTMM) \cite{Chinaei2010} have also been evaluated. 
Goal-driven approach can create both pro-active and reactive chatbots, while non-goal-driven approach creates reactive chatbots. In addition, they can serve as a tool to goal-driven systems as in \cite{Bordes2016}. That is, when trained on corpora of a goal-driven 
system, non-goal-driven systems can be used to simulate user interaction to then train goal-driven models.

\subsection{Types of Context Reasoning}

A dialogue system may support the context reasoning or not. Context reasoning is necessary in many occasions. For instance, when partial information is provided the chatbot needs to be able to interact one or more turns in order to get the complete information in order to be able to properly answer. In \cite{Higashinaka2015}, the  authors present a taxonomy of errors in conversational systems. The ones regarding context-level errors are the ones that are perceived as the top-10 confusing and they are mainly divided into the following:

\begin{itemize}
\item Excess/lack of proposition: the utterance does not provide any new proposition to the discourse context or provides excessive information than required.
\item Contradiction: the utterance contains propositions that contradict what has been said by the system or by the user.
\item Non-relevant topic: the topic of the utterance is irrelevant to the current context such as when the system suddenly jumps to some other topic triggered by some particular word in the previous user utterance.
\item Unclear relation: although the utterance might relate to the previous user utterance, its relation to the current topic is unclear. 
\item Topic switch error: the utterance displays the fact that the system missed the switch in topic by the user, continuing with the previous topic.
\end{itemize}

\noindent \textit{Rule-oriented} \\

In the state of the art most of the proposed approaches for context reasoning lies on rules using logics and knowledge bases as described in the Rule-oriented architecture sub-section. Given a set of facts extracted from the dialogue history and encoded in, for instance, FOL statements, a queries can be posed to the inference engine and produce answers. For instance, see the example in Table \ref{tab:example}. The sentences were extracted from \cite{Sukhbaatar2015} (which does not use a rule-oriented approach), and the first five statements are their respective facts. The system then apply context reasoning for the query \textit{Q: Where is the apple}.\\

\begin{table}[h]
  \caption{Example of sequence of utterances and FOL statements}
  \centering
    \begin{tabular}{|l|l|}
    \hline
\textbf{Utterance} & \textbf{FOL statement} \\\hline\hline
Sam walks into the kitchen. & $isAt(Sam, kitchen)$ \\\hline
Sam picks up an apple. & $pickUp(Sam, apple)$ \\\hline
Sam walks into the bedroom. & $isAt(Sam, bedroom)$ \\\hline
Sam drops the apple. &  $\forall{x} (\forall{y} ( \forall{w}( drops(x, y) \wedge isAt(x, w) \to isAt(y, w) ) ) ) $ \\\hline
Q: Where is the apple? &   \\\hline
A: Bedroom & isAt(apple, bedroom) \\\hline
  \end{tabular}
   \vspace{-1\baselineskip}
  \label{tab:example}
\end{table}

If statements above are received on the order present in Table  \ref{tab:example}, if the query \textit{Q: Where is the apple} is sent, the inference engine will produce the answer \textit{A: Bedroom} (i.e., the statement $isAt(apple, bedroom)$ is found by the model and returned as True). 

Nowadays, the most common way to store knowledge bases is on triple stores, or RDF (Resource Description Framework)\footnote{The Resource Description Framework (RDF) is a standard model for expressing graph data for the World Wide Web.} stores. A triple store is a knowledge base for the storage and retrieval of triples through semantic queries. A triple is a data entity composed of subject-predicate-object, like "Sam is at the kitchen" or "The apple is with Sam", for instance. A query language is needed for storing and retrieving data from a triple store. While SPARQL is a RDF query language, Rya\footnote{https://rya.apache.org/} is an open source scalable RDF triple store built on top of Apache Accumulo\footnote{https://accumulo.apache.org/}. Originally developed by the Laboratory for Telecommunication Sciences and US Naval Academy, Rya is currently being used by a number of american government agencies for storing, inferencing, and querying large amounts of RDF data. 

A SPARQL query has a SQL-like syntax for finding triples matching specific patterns. For instance, see the query below. It retrieves all the people that works at IBM and lives in New York: \\

\noindent \texttt{SELECT ?people } \\
\noindent \texttt{WHERE \{} \\
\noindent \texttt{?people <worksAt> <IBM> . } \\
\noindent \texttt{?people <livesIn> <New York>. }\\
\noindent \texttt{\}} \\

Since triple stores can become huge, Rya provides three triple table index \cite{Meier2015} to help speeding up queries: \\

\noindent \textit{SPO}: subject, predicate, object\\
\noindent \textit{POS}: predicate, object, subject\\
\noindent \textit{OSP}: object, subject, predicate\\

While Rya is an example of an optimized triple store, a rule-oriented chatbot can make use of Rya or any triple store and can call the semantic search engine in order to inference and generate proper answers. \\

\noindent \textit{Data-oriented} \\

Recent papers have used neural networks to predict the next utterance on non-goal-driven systems considering the context, for instance with 
Memory Networks \cite{Weston2015}. In this work \cite{Sukhbaatar2015}, for example the authors were able to generate answers for dialogue like below: \\

\noindent \texttt{Sam walks into the kitchen.}\\
\noindent \texttt{Sam picks up an apple.}\\
\noindent \texttt{Sam walks into the bedroom.}\\   
\noindent \texttt{Sam drops the apple.}\\
\noindent \texttt{Q: Where is the apple?}\\
\noindent \texttt{A: Bedroom}\\

Sukhbaatar's model represents the sentence as a vector in a way that the order of the words matter, and the model encodes the temporal context enhancing the memory vector with a matrix that contains the temporal information. During the execution phase, Sukhbaatar's model takes a discrete set of inputs $x_1, ..., x_n$ that are to be stored in the memory, a query $q$, and outputs an answer $a$. Each of the $x_i$, $q$, and $a$ contains symbols coming from a dictionary with $V$ words. The model writes all $x$ to the memory up to a fixed buffer size, and then finds a continuous representation for the $x$ and $q$. The continuous representation is then processed via multiple computational steps to output $a$. This allows back propagation of the error signal through multiple memory accesses back to the input during training. Sukhbaatar's also presents the state of the art of recent efforts that have explored ways to capture dialogue context, treated as long-term structure within sequences, using RNNs or LSTM-based models. The problem of this approach is that it is has not been tested for goal-oriented systems. In addition, it works with a set of sentences but not necessary from multi-party bots.

\subsection{Platforms}
 
Regarding current platforms to support the development of conversational systems, we can categorize them into three types:  platforms for plugging chatbots, for creating chatbots and for creating service chatbots. The platforms for plugging chatbots provide tools for integrating them another system, like Slack\footnote{www.slack.com}. The chatbots need to receive and send messages in a specific way, which depends on the API and there is no support for actually helping on building chatbots behavior with natural language understanding. The platforms for creating chatbots mainly provide tools for adding and training intentions together with dialogue flow specification and some entities extraction, with no reasoning support. Once the models are trained and the dialogue flow specified, the chatbots are able to reply to the received intention. The platforms for creating service chatbots provide the same functionalities as the last one and also provide support for defining actions to be executed by the chatbots when they are answering to an utterance. Table \ref{tab:platforms} summarizes current platforms on the market accordingly to these categories. There is a lack on platforms that allow to create chatbots that can be coordinated in a multiparty chat with governance or mediation.

\begin{table}[h]%
\caption{Platforms for Building Chatbots}
\centering
\begin{tabular}{|p{2.6cm}|p{2cm}|p{2cm}|p{2.2cm}|p{2.5cm}|}
\hline
 & \textbf{Plugging Bots}  & \textbf{Creating Bots}  & \textbf{Creating Service Bots} & \textbf{Coordination Control}    \\ \hline  
IBM (Watson) & \ding{109} &   \ding{108} &  \ding{119}  & \ding{109}  \\ \hline
Pandora   & \ding{109}  & \ding{108}  &  \ding{108} & \ding{109}  \\ \hline
Facebook (Wit.ai) & \ding{109}  &  \ding{108}  &  \ding{108}  &  \ding{109} \\ \hline
Microsoft (LUIS) &  \ding{109} &   \ding{108} &  \ding{109} & \ding{109}  \\ \hline
Google Hangout & \ding{108} & \ding{109}  &  \ding{109}  & \ding{109}  \\ \hline
Slack &   \ding{108} & \ding{109}  & \ding{109}  & \ding{109}  \\ \hline
WeChat &   \ding{108}  &   \ding{108}  &   \ding{108} &  \ding{109} \\ \hline
Kik &  \ding{108}  & \ding{109}  &  \ding{109} &  \ding{109} \\ \hline
\end{tabular}
\label{tab:platforms}
\vspace{-1\baselineskip}
\end{table}

\section{A Conceptual Architecture for Multiparty-Aware Chatbots}
\label{sec:arch}

\begin{figure}[h]
\centerline{\includegraphics[width=\textwidth]{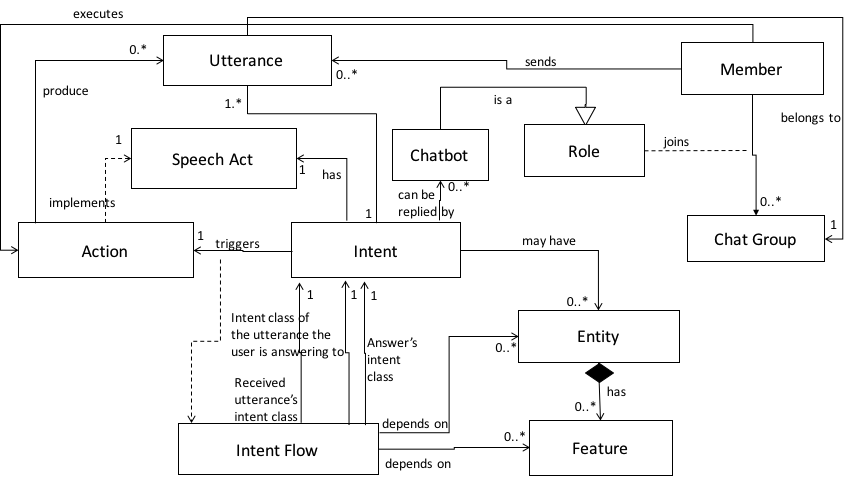}}
\caption{Chatbot Knowledge's Conceptual Model in a \gls{MPCS}}
\label{fig:conceptual-model}
\end{figure}

In this section the conceptual architecture for creating a hybrid rule and machine learning-based \gls{MPCS} is presented. The \gls{MPCS} is defined by the the entities and relationships illustrated in Fig. \ref{fig:conceptual-model} which represents the chatbot's knowledge. A \texttt{Chat Group} contains several \texttt{Members} that join the group with a \texttt{Role}. The role may constrain the behavior of the member in the group. \texttt{Chatbot} is a type of \texttt{Role}, to differentiate from persons that may also join with different roles. For instance, a person may assume the role of the owner of the group, or someone that was invited by the owner, or a domain role like an expert, teacher or other.

When a \texttt{Member} joins the \texttt{Chat Group}, it/he/she can send \texttt{Utterances}. The \texttt{Member} then classifies each \texttt{Utterance} with an \texttt{Intent} which has a \texttt{Speech Act}\footnote{Or dialogue act, or communicative act, of performative, depending on the research community.}. The \texttt{Intent} class, \texttt{Speech Act} class and the \texttt{Intent Flow} trigger the \texttt{Action} class to be executed by the \texttt{Member} that is a \texttt{Chatbot}. The \texttt{Chatbots} associated to the \texttt{Intention} are the only ones that know how to answer to it by executing \texttt{Actions}. The \texttt{Action}, which implements one \texttt{Speech Act}, produces answers which are \texttt{Utterances}, so, for instance, the \texttt{Get\_News} action produces an \texttt{Utterance} for which \texttt{Intention}'s speech act is \texttt{Inform\_News}. The \texttt{Intent Flow} holds the intent's class conversation graph which maps the dialog state as a decision tree. The answer's intention class is mapped in the \texttt{Intent Flow} as a directed graph G defined as following:

\begin{equation}
G=(V,R) | (x,y) = ( <I_u, I_r>,  I_a)
\label{eq:graph}
\end{equation}

From the graph definitions, $V$ is for \textit{vertices} and $R$ is for \textit{relations}, which are the arrows in the graph. And in Equation \ref{eq:graph}:
\noindent $V$ is the set of intentions pairs,

\noindent  $R$ is the set of paths to navigate through the intentions, 

\noindent $<I_u, I_r>$ is the arrow's head, and

\noindent  $I_a$ is the arrow's tail.  \\

This arrow represents a turn from an utterance with $I_u$ intention class which is replying to an utterance with $I_r$ intention class to the state which an utterance with $I_r$ intention's class  is sent.

$I_a$ is the intention class of the answer to be provided to the received $I_u$ intention class.

In addition, each intent's class may refer to many \texttt{Entities} which, in turn, may be associated to several \texttt{Features}. For instance, the utterance \\

 \textit{"I would like to invest USD10,000 in Savings Account for 2 years"} \\
 
 contains one entity -- the Savings Account's investment option -- and two features -- money (USD10,000) and period of time (2 years). The \texttt{Intent Flow} may need this information to choose the next node which will give the next answer. Therefore, if the example is changed a little, like\\
  
  \textit{"I would like to invest in Savings Account"}, \\
  
  $I_a$ is constrained by the "Savings Account" entity which requires the two aforementioned features. Hence, a possible answer by one \texttt{Member} of the group would be \\
  
  \textit{"Sure, I can simulate for you, what would be the initial amount and the period of time of the investment?"} \\

With these conceptual model's elements, a \gls{MPCS} system can be built with multiple chatbots. Next subsection further describes the components workflow.

\subsection{Workflow}

\begin{figure}[hb!]
\centerline{\includegraphics[width=\textwidth]{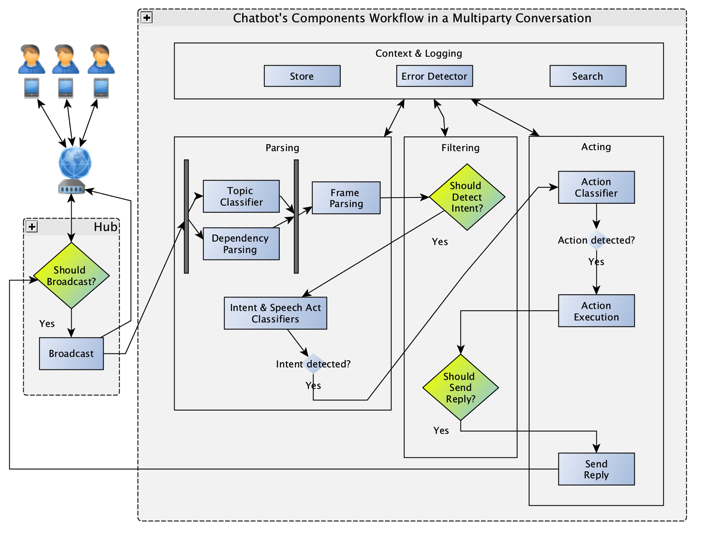}}
\caption{Workflow}
\label{fig:workflow}
\end{figure}

Figure \ref{fig:workflow} illustrates from the moment that an utterance is sent in a chat group to the moment a reply is generated in the same chat group, if the case. One or more person may be in the chat, while one or more chatbots too. There is a \texttt{Hub} that is responsible for broadcasting the messages to every \texttt{Member} in the group, if the case. The flow starts when a \texttt{Member} sends the utterance which goes to the \texttt{Hub} and, if allowed, is broadcasted. Many or none interactions norms can be enforced at this level depending on the application. Herein, a norm can be a prohibition, obligation or permission to send an utterance in the chat group.

Once the utterance is broadcasted, a chatbot needs to handle the utterance. In order to properly handle it, the chatbot parses the utterance with several parsers in the \texttt{Parsing phase}: a \texttt{Topic Classifier}, the \texttt{Dependency Parsing}, which includes Part-of-Speech tags and semantics tags, and any other that can extract metadata from the utterance useful for the reasoning. All these metadata, together with more criteria, may be used in the \texttt{Frame parsing} which is useful for context reasoning. All knowledge generated in this phase can be stored in the \texttt{Context}. Then, the \texttt{Intent Classifier} tries to detect the intent class of the utterance. If detected, the \texttt{Speech Act} is also retrieved.  And an \texttt{Event Detector} can also check if there is any dialog inconsistency during this phase.

After that, the \texttt{Filtering phase} receives the object containing the utterance, the detected intent, and all metadata extracted so far and decides if an action should be performed to reply to the utterance. If yes, it is sent to the \texttt{Acting phase} which performs several steps. First the \texttt{Action Classifier} tries to detect the action to be performed. If detected, the action is executed. At this step, many substeps may be performed, like searching for an information, computing maths, or generating information to create the answer. All of this may require a search in the \texttt{Context} and also may activate the \texttt{Error Detector} component to check if the dialog did not run into a wrong state. After the answer is generated, the \texttt{Filtering phase} is activated again to check if the reply should be really sent. If so, it is sent to the \texttt{Hub} which, again may check if it can be broadcasted before actually doing it.

\subsubsection{Topic Classifier}

The topic classifier is domain-dependent and is not mandatory. However, the chatbot can better react when the intent or action is not detected, which means that it does not know how to answer. Many reasons might explain this situation: the set of intents might be incomplete, the action might not have produced the proper behavior, misunderstanding might happen, or the chatbot was not designed to reply to a particular topic. In all cases, it must be able to produce a proper reply, if needed. Because this might happen throughout the workflow, the sooner that information is available, the better the chatbot reacts. Therefore it is one of the first executions of the flow.

\subsubsection{Dependency Parsing}

Dependency is the notion that linguistic units, e.g. words, are connected to each other by directed links. The (finite) verb is taken to be the structural center of clause structure. All other syntactic units (words) are either directly or indirectly connected to the verb in terms of the directed links, which are called dependencies. It is a one-to-one correspondence: for every element (e.g. word or morph) in the sentence, there is exactly one node in the structure of that sentence that corresponds to that element. The result of this one-to-one correspondence is that dependency grammars are word (or morph) grammars. All that exist are the elements and the dependencies that connect the elements into a structure. Dependency grammar (DG) is a class of modern syntactic theories that are all based on the dependency relation.

Semantic dependencies are understood in terms of predicates and their arguments. Morphological dependencies obtain between words or parts of words. To facilitate future research in unsupervised induction of syntactic structure and to standardize best-practices, a tagset that consists of twelve universal part-of-speech categories was proposed \cite{Petrov11auniversal}.

Dependency parsers have to cope with a high degree of ambiguity and nondeterminism which let to different techniques than the ones used for parsing well-defined formal languages. Currently the mainstream approach uses algorithms that derive a potentially very large set of analyses in parallel and when disambiguation is required, this approach can be coupled with a statistical model for parse selection that ranks competing analyses with respect to plausibility \cite{Nivre2008}.

Below we present an example of a dependency tree for the utterance: \\ \\ \textit{"I want to invest 10 thousands"}:

\lstset{
    string=[s]{"}{"},
    stringstyle=\color{blue},
    comment=[l]{:},
    commentstyle=\color{black},
}
\begin{lstlisting}
   "tree": {
      "want VERB ROOT": {
        "I PRON nsubj": {},
        "to ADP mark": {}, 
        "invest VERB nmod": {
           "thousands NOUN nmod": { 
               "10 NUM nummod": {} 
           } 
        }
      }
  
\end{lstlisting}

The coarse-grained part-of-speech tags, or morphological dependencies (\texttt{VERB}, \texttt{PRON}, \texttt{ADP}, \texttt{NOUN} and \texttt{NUM}) encode basic grammatical categories and the grammatical relationships (\texttt{nsubjs}, \texttt{nmod}, \texttt{nummod}) are defined in the Universal Dependencies project\footnote{http://universaldependencies.org/} \cite{Petrov11auniversal}.

\subsubsection{Frame Parsing} 

In this module, the dependency tree generated is used together with a set of rules to extract information that is saved in the context using the Frame-based approach. This approach fills the slots of the \textit{frame} with the extracted values from the dialogue. Frames are like forms and slots are like fields. Using the knowledge's conceptual model, the fields are represented by the elements \texttt{Entities} and \texttt{Features}. In the dependency tree example, the entity would be the implicit concept: the investment option, and the feature is the implicit concept: initial amount -- \texttt{10 thousands}. Since the goal is \texttt{to invest}, and there are more entities needed for that (i.e., fields to be filled), the next node in the Intention Flow tree would return an utterance which asks the user the time of investment, if he/she has not provided yet.

This module could be implemented using different approaches according to the domain, but tree search algorithms will be necessary for doing the tree parsing.

\subsubsection{Intent Classifier} 

The \texttt{Intent Classifier} component aims at recognizing not only the \texttt{Intent} but the \textit{goal} of the utterance sent by a \texttt{Member}, so it can properly react. The development of an intent classifier needs to deal with the following steps: \\

\noindent i) the creation of dataset of intents, to train the classification algorithm;  \\
\noindent ii) the design of a classification algorithm that provides a reasonable level of accuracy;\\
\noindent iii) the creation of dataset of trees of intents, the same as defined in i) and which maps the goals;  \\
\noindent iv) the design of a plan-graph search algorithm that maps the goal's state to a node in the graph;\\

There are several approaches to create training sets for dialogues: from an incremental approach to crowdsourcing. In the incremental approach, the Wizard of Oz method can be applied to a set of potential users of the system, and from this study, a set of questions that the users asked posted to the `fake' system can be collected. These questions have to be manually classified into a set of intent classes, and used to train the first version of the system. Next, this set has to be increased both in terms of number of classes and samples per class.

\subsubsection{Speech Act Classifier}

The \texttt{Speech Act Classifier} can be implemented with many speech act classes as needed by the application. The more classes, the more flexible the chatbot is. It can be built based on dictionaries, or a machine learning-based classifier can be trained. In the table below we present the main and more general speech act classes \cite{FIPA} used in the Chatbots with examples to differentiate one from another:

\begin{table}[h]%
\caption{Main Generic Speech Acts Classes}
\centering
\begin{tabular}{|p{3.5cm}|p{8cm}|}
\hline
    \textbf{Speech Act}   &   \textbf{Example}        \\ \hline
GREETINGS &   \textit{Hello}   \\ \hline
THANK &   \textit{Thank you very much}   \\ \hline
INFORM &   \textit{The stock market is launched}   \\ \hline
QUERY &    \textit{Is the stock market launched?}   \\ \hline
CFP &   \textit{Can someone launch the stock market?}   \\ \hline
REQUEST &  \textit{Launch the stock market}  \\ \hline
AGREE &  \textit{OK, I will launch the stock market}  \\ \hline
REFUSE &  \textit{I will not launch the stock market} \\ \hline
FAILURE &  \textit{I can not launch the stock market}  \\ \hline
PROPOSE &  \textit{I can launch the stock market for X dollars} \\ \hline
SUBSCRIBE & \textit{I would like to know when someone launches the stock market}  \\ \hline
NOT-UNDERSTOOD &  \textit{Stock market? Which stock market?}  \\ \hline
BYE &   \textit{See you}   \\ \hline
\end{tabular}
 \vspace{-1\baselineskip}
 \label{tab:speechacts}
\end{table}%

\subsubsection{Action Classifier}

There are at least as many \texttt{Action} classes as \texttt{Speech Act} classes, since the action is the realization of a speech act. The domain specific classes, like \texttt{"Inform\_News"} or \texttt{"Inform\_Factoids"}, enhance the capabilities of answering of a chatbot.

The \texttt{Action Classifier} can be defined as a multi-class classifier with the tuple

\begin{equation}
\label{eq:actionclasseq}
A=<I_a, S_a, E, F>
\end{equation}

where $I_a$ is the intent of the answer defined in (\ref{eq:graph}), $S_a$ is the speech act of the answer, $E$ and $F$ are the sets of entities and features needed to produce the answer, if needed, respectively.

\subsubsection{Action Execution} 

This component is responsible for implementing the behavior of the \texttt{Action} class. Basic behaviors may exist and be shared among different chatbots, like the ones that implement the greetings, thanks or not understood. Although they can be generic, they can also be personalized to differentiate the bot from one another and also to make it more "real". Other cases like to inform, to send a query, to send a proposal, they are all domain-dependent and may require specific implementations.

\begin{figure}[h!]
\centerline{\includegraphics[width=0.6\textwidth]{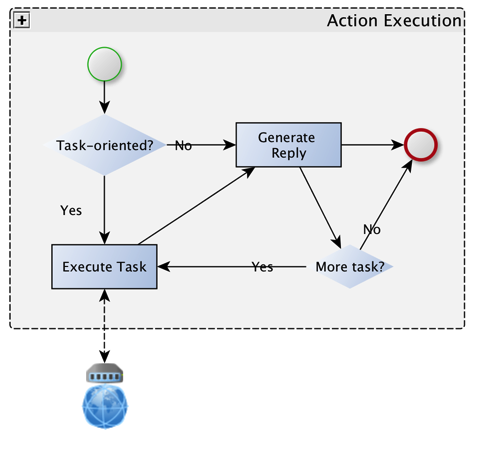}}
\caption{Action Execution}
\label{fig:action-execution}
\end{figure}

Anyway, figure \ref{fig:action-execution} shows at the high level the generic workflow. If action class detected is task-oriented, the system will implement the execution of the task, say to guide a car, to move a robot's arm, or to compute the return of investments. The execution might need to access an external service in the Internet in order to complete the task, like getting the inflation rate, or the interest rate, or to get information about the environment, or any external factor. During the execution or after it is finished, the utterance is generated as a reply and, if no more tasks are needed, the action execution is finished.

In the case of coordination of chatbots, one or more chatbots with the role of mediator may exist in the chat group and, at this step, it is able to invite one or more chatbots to the chat group and it is also able to redirect the utterances, if the case.

The proposed architecture addresses the challenges as the following:

\begin{enumerate}
\item \textit{\textbf{What} is the message/utterance about?} solved by the \textbf{\textit{Parsing phase}}; 
\item \textit{\textbf{Who} should reply to the utterance?}  solved by the \textbf{\textit{Filtering phase}} and may be enforced by the \textbf{\textit{Hub}}; 
\item \textit{\textbf{How} the reply should be built/generated?} solved by the \textbf{\textit{Acting phase}};
\item \textit{\textbf{When} should the reply be sent?} may be solved by the \textbf{\textit{Acting phase}} or the \textbf{\textit{Filtering phase}}, and may be enforced by the \textbf{\textit{Hub}}; 
\end{enumerate}

And \texttt{Context and Logging} module is used throughout all phases.

\section{Architecture Implementation and Evaluation}
\label{sec:eval}

This section presents one implementation of the conceptual architecture presented in last section. After many refactorings, a framework called SABIA (Speech-Act-Based Intelligent Agents Framework) has been developed and CognIA (Cognitive Investment Advisor) application has been developed as an instantiation of SABIA framework. We present then the accuracy and some automated tests of this implementation.

\subsection{Speech-Act-based Intelligent Agents Framework}

SABIA was developed on top of Akka middleware\footnote{http://akka.io}. Akka is a toolkit and runtime that implements the Actor Model on the JVM. Akka's features, like concurrency, distributed computing, resilience, and message-passing were inspired by Erlang's actor model \cite{Armstrong2010} \cite{Hewitt1973}. The actor model is a mathematical model of concurrent computation that treats "actors" as the universal primitives of concurrent computation. In response to a message that it receives, an actor can: make local decisions, create more actors, send more messages, and determine how to respond to the next received message. Actors may modify private state, but can only affect each other through messages (avoiding the need for any locks). Akka middleware manages the actors life cycle and actors look up by theirs name, locally or remotely.

We implemented each \texttt{Member} of the \texttt{Chat Group} as an \texttt{Actor} by extending the \texttt{UntypedActor} class of Akka middleware. Yet, we created and implemented the \texttt{SabiaActorSystem} as a singleton (i.e., a single instance of it exists in the system) \cite{Gamma1994} that has a reference to Akka's \texttt{ActorSystem}. During \texttt{SabiaActorSystem}'s initialization, all parsers that consume too much memory during their initialization to load models are instantiated as singletons. In this way, we save time on their calls during the runtime.  Moreover, all chat group management, like to join or leave the group, or to broadcast or filter a message at the \texttt{Hub} level is implemented in SABIA through the \texttt{Chat Group} behavior.

\subsubsection{Dependency Parsing}

This is implemented in SABIA as a singleton that is initialized during the \texttt{SabiaActorSystem} initialization with the URL of the service that implements the dependency parsing and is used on each utterance's arrival through the execution of the \texttt{tagUtterance} method. The service must retrieve a \textit{JSON Object} with the dependency tree which is then parsed using depth-first search.

\subsubsection{Frame Parsing} 

SABIA does not support invariants for frame parsing. We are leaving this task to the instantiated application.

\subsubsection{Intent Classifier} 

There are two intent classifiers that can be loaded with trained models in order to be ready to be used at runtime: the 1-nearest-neighbor (1NN) and the SVM-based classifier.

\subsubsection{Action Classifier}

SABIA implements the \texttt{Action Classifier} assuming that the application uses a relational database with a data schema that implements the conceptual model presented in Figure \ref{fig:conceptual-model}. Then the invariants parts that use SQL are already present and the application only needs to implement the database connection and follow the required data schema.

\subsubsection{Action Execution} 

SABIA provides partial implemented behavior for the \texttt{Action} through the \texttt{Template method} design pattern \cite{Gamma1994}, which implements the invariants parts of the action execution and leaves placeholders for customization.

\begin{table}[t!]
\caption{MultiParty Conversation with Mixed initiative}
\centering
\begin{tabular}{|p{0.4cm}|p{3.3cm}|p{8.2cm}|}
\hline
&  \textbf{Turn} & \textbf{Utterance} \\\hline\hline
1 & User & \textit{"I have \$30,000 USD, where should I invest it?"} \\\hline
2 & Mediator chatbot &  \textit{"Well, for how long could you keep the money invested?"} \\\hline
3 & User & \textit{"Say, for 2 years."} \\\hline
4 & Mediator chatbot & \textit{"All right, then, considering the amount and the time period, why don't you simulate this investment in a savings account?"} \\\hline
5 & User &  \textit{"Sure! I would love to."} \\\hline
6 & Mediator chatbot & \textit{"Ok, I will invite the savings account to this group."} \\\hline
7 & Savings account chatbot & $<<$ Savings account chatbot joins the group $>>$ \\\hline
8 & Mediator chatbot & \textit{"Hi Savings Account expert, could you please simulate the return of investment of \$30,000 in 2 years?"} \\\hline
9 & Savings account chatbot & \textit{"Sure, just a minute..."} \\\hline
10 & Savings account chatbot & \textit{"Well, at the end, one would have \$32,500 USD."} \\\hline
11 & Mediator chatbot & \textit{"Thank you. Well, it seems a good idea given the economy right now."} \\\hline
12 & User & \textit{"Thank you all."} \\\hline
13 & Savings account chatbot & \textit{"You're welcome."}  \\\hline
14 & Mediator chatbot & \textit{"No problem. Let me know if I can help you with something else."} \\\hline
\end{tabular}
 \vspace{-1\baselineskip}
 \label{tab:multiparty-cognia}
\end{table}

\subsection{CognIA: A Cognitive Investment Advisor}

We developed CognIA, which is an instantiation of Sabia framework. A conversation is composed of a group chat that can contain multiple users and multiple chatbots. This example, in particular, has a mediator that can help users on financial matters, more specifically on investment options.  For example, consider the following dialogue in the table below: \\

The Table \ref{tab:multiparty-cognia} shows an example that uses the mixed-initiative dialogue strategy, and a dialogue mediator to provide coordination control.  In this example of an application, there are many types of intentions that should be answered: Q\&A (question and answer) about definitions, investment options, and about the current finance indexes, simulation of investments, which is task-oriented and requires computation, and opinions, which can be highly subjective. \\

In Table \ref{tab:rules}, we present the interaction norms that were needed in Cognia. The \texttt{Trigger} column describes the event that triggers the \texttt{Behavior} specified in the third column. The \texttt{Pre-Conditions} column specifies what must happen in order to start the behavior execution. So, for instance, line 2, when the user sends an utterance in the chat group, an event is triggered \textbf{and}, if the utterance's topic is CDB (Certificate of Deposit which is a fixed rate investment) or if it is about the Savings Account investment option \textbf{and} the speech act is not \texttt{Query\_Calculation} \textbf{and} the CDB and Savings Account members are not in the chat, \textbf{then} the behavior is activated. The bot members that implement these behaviors are called \textit{cdbguru} and \textit{poupancaguru}. Therefore these names are used when there is a mention. 

Note that these interactions norms are not explicitly defined as obligations, permissions, and prohibitions. They are implict from the behavior described. During this implementation, we did not worry about explicitly defining the norms, because the goal was to evaluate the overall architecture, not to enhance the state of the art on norms specification for conversational systems. In addition, CognIA has only the presented interaction norms defined in  Table \ref{tab:rules}, which is a very small set that that does not required model checking or verification of conflicts.

\begin{center}
\centering
\begin{longtable}{|p{2cm}|p{5.0cm}|p{5.0cm}|}%
\caption[Cognia Interaction Norms]{Cognia Interaction Norms}  \label{tab:rules}\\
\hline
\textbf{Trigger} & \textbf{Pre-Conditions} & \textbf{Behavior} \\\hline\hline
On group chat creation & Cognia chatbot is available & Cognia chatbot joins the chat with the \textit{mediator} role and user joins the chat with the \textit{owner\_user} role \\\hline
On utterance sent by user & Utterance's topic is CDB (\textit{cdbguru}) \textbf{or} Savings Account (\textit{poupancaguru}) \textbf{and} speech act is not \textit{Query\_Calculation} \textbf{and} they are not in the chat & Cognia invites experts to the chat and repeats the utterance to them  \\\hline
On utterance sent by user & Utterance's topic is CDB (\textit{cdbguru}) \textbf{or} Savings Account (\textit{poupancaguru}) \textbf{and} speech act is not \textit{Query\_Calculation} \textbf{and} they are in the chat & Cognia \textbf{waits for while} and \textit{cdbguru} or \textit{poupancaguru} respectively handles the utterance. If they don't understand, they \textbf{don't reply} \\\hline
On utterance sent by the experts & If Cognia is waiting for them \textbf{and} has received both replies & Cognia does not wait anymore \\\hline
On utterance sent & Utterance mentions \textit{cdbguru} \textbf{or} \textit{poupancaguru} & \textit{cdbguru} or \textit{poupancaguru} respectively handles the utterance \\\hline
On utterance sent & Utterance mentions \textit{cdbguru} \textbf{or} \textit{poupancaguru} \textbf{and} they don't reply after a while \textbf{and} speech act is \textit{Query\_Calculation}  & Cognia sends \textit{I can only chat about investments...} \\\hline
On utterance sent & Utterance mentions \textit{cdbguru} \textbf{or} \textit{poupancaguru} \textbf{and} they don't reply after while \textbf{and} speech act is not \textit{Query\_Calculation}  & Cognia sends \textit{I didn't understand} \\\hline
On utterance sent & Utterance's speech act is \textit{Query\_Calculation} \textbf{and} period or initial amount of investment were \textbf{not} specified & Cognia asks the user the missing information  \\\hline
On utterance sent & Utterance's speech act is \textit{Query\_Calculation} \textbf{and} period and initial amount of investment were specified \textbf{and} the experts are \textbf{not} in the chat & Cognia invites experts to the chat and repeats the utterance to them  \\\hline
On utterance sent & Utterance's speech act is \textit{Query\_Calculation} \textbf{and} period and initial amount of investment were specified \textbf{and} the experts are in the chat & Cognia repeats the utterance to experts  \\\hline
On utterance sent & Utterance's speech act is \textit{Query\_Calculation}  & Cognia extracts variables and saves the context \\\hline
On utterance sent & Utterance's speech act is \textit{Query\_Calculation}  \textbf{and} the experts are in the chat  \textbf{and} the experts are mentioned & Experts extract information, save in the context, compute calculation and send information \\\hline
On utterance sent & Utterance's speech act is \textit{Inform\_Calculation} \textbf{and} Cognia received all replies & Cognia compares the results and inform comparison \\\hline
On utterance sent & Utterance mentions a chatbot but has \textbf{no} other text & The chatbot replies \textit{How can I help you?} \\\hline
On utterance sent & Utterance is not understood \textbf{and} speech act is \textit{Question} & The chatbot replies \textit{I don't know... I can only talk about topic X} \\\hline
On utterance sent & Utterance is not understood \textbf{and} speech act is \textbf{not} \textit{Question} & The chatbot replies \textit{I didn't understand} \\\hline
On utterance sent & Utterance's speech act is one of \{ \textit{Greetings, Thank, Bye} \}  & All chatbots reply to utterance \\\hline
On group chat end &  & All chatbots leave the chat, and the date and time of the end of chat is registered \\\hline
\end{longtable}
\end{center}%

\subsubsection{Instantiating SABIA to develop CognIA}

We instantiated SABIA to develop CognIA as follows: the \texttt{Mediator}, \texttt{Savings Account}, \texttt{CDB} and \texttt{User Actors} are the \texttt{Members} of the \texttt{Chat Group}. The \texttt{Hub} was implemented using two servers: \texttt{Socket.io} and \texttt{Node.JS} which is a socket client of the \texttt{Socket.io} server. The CognIA system has also one \texttt{Socket Client} for receiving the broadcast and forwarding to the \texttt{Group Chat Manager}. The former will actually do the broadcast to every member after enforcing the norms that applies specified in Table \ref{tab:rules}. Each \texttt{Member} will behave according to this table too. For each user of the chat group, on a mobile or a desktop, there is its corresponding actor represented by the \texttt{User Actor} in the figure. Its main job is to receive Akka's broadcast and forward to the Socket.io server, so it can be finally propagated to the users.

All the intents, actions, factual answers, context and logging data are saved in DashDB (a relational Database-as-a-Service system). When an answer is not retrieved, a service which executes the module \texttt{Search Finance on Social Media} on a separate server is called. This service was implemented with the assumption that finance experts post relevant questions and answers on social media. Further details are explained in the \textit{Action execution} sub-section.

\begin{figure}[h]
\centerline{\includegraphics[width=\textwidth]{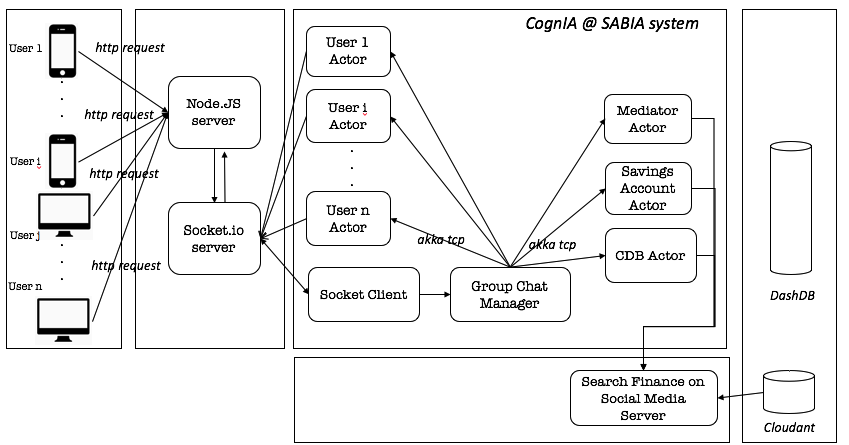}}
\caption{Cognia Deployment View}
\label{fig:cognia}
\end{figure}


\subsubsection{Topic Classifier}

We built a small dictionary-based topic classifier to identify if an utterance refers to finance or not, and if it refers to the two investment options (CDB or Savings Account) or not.

\subsubsection{Dependency Parsing}

The dependency parsing is extremely important for computing the return of investment when the user sends an utterance with this intention. Our first implementation used regular expressions which led to a very fragile approach. Then we used a TensorFlow implementation\footnote{TensorFlow is an open-source software library for numerical computation using data flow graphs. Nodes in the graph represent mathematical operations, while the graph edges represent the multidimensional data arrays (tensors) communicated between them. URL: https://www.tensorflow.org/} \cite{tensorflow2015} of a SyntaxNet model for Portuguese and used it to generate the dependency parse trees of the utterances. The SyntaxNet model is a feed-forward neural network that operates on a task-specific transition system and achieves the state-of-the-art on part-of-speech tagging, dependency parsing and sentence compression results \cite{Andor2016}. Below we present output\footnote{Converted from Brazilian Portuguese to English.} of the service for the utterance: \\ \\ \textit{"I want to invest 10 thousands in 40 months"}:

\lstset{
    string=[s]{"}{"},
    stringstyle=\color{blue},
    comment=[l]{:},
    commentstyle=\color{black},
}
\begin{lstlisting}
{ "original": "I would like to invest 10 thousands in 40 months", 
  "start_pos": [
    23, 
    32], 
  "end_pos": [
    27, 
    33], 
  "digits": [
    10000, 
    40], 
  "converted": "I would like to invest 10000 in 40 months", 
  "tree": {
    "like VERB ROOT": {
        "I PRON nsubj": {}, 
        "would MD aux":{
            "invest VERB xcomp":{
                 "to TO aux": {}, 
                 "10000 NUM dobj": {}, 
                 "in IN prep": {
                      "months NOUN pobj":{
                          "40 NUM num": {}}}}}}}
\end{lstlisting} 

The service returns a \texttt{JSON Object} containing six fields: \texttt{original, start\_pos, end\_pos, digits, converted} and \texttt{tree}. The \texttt{original} field contains the original utterance sent to the service. The \texttt{converted} field contains the utterance replaced with decimal numbers, if the case (for instance, \texttt{"10 thousands"} was converted to \texttt{"10000"} and replaced in the utterance). The \texttt{start\_pos} and \texttt{end\_pos} are arrays that contain the start and end char positions of the numbers in the converted utterance. While the \texttt{tree} contains the dependency parse tree for the converted utterance.

\subsubsection{Frame Parsing} 

Given the dependency tree, we implemented the frame parsing which first extracts the entities and features from the utterance and saves them in the context. Then, it replaces the extracted entities and features for reserved characters. 

\begin{algorithm}
  \caption{extract\_period\_of\_investment (utteranceTree)}\label{getperiod}
  \begin{algorithmic}[1]
    \State \begin{varwidth}[t]{\linewidth}
    numbersNodes~$\gets$~utteranceTree.getNumbersNodes();
      \end{varwidth}
    \State \begin{varwidth}[t]{\linewidth}
    \textbf{foreach}(numberNode \textbf{in} numbersNodes)  \textbf{do}
      \end{varwidth}
    \State \begin{varwidth}[t]{\linewidth}
    \hskip\algorithmicindent parentsOfNumbersNode~$\gets$~numbersNode.getParents()
      \end{varwidth}
    \State \begin{varwidth}[t]{\linewidth}
    \hskip\algorithmicindent \textbf{foreach}(parent \textbf{in} parentsOfNumbersNodes)  \textbf{do}
      \end{varwidth}
     \State \begin{varwidth}[t]{\linewidth}
     \hskip\algorithmicindent \hskip\algorithmicindent \textbf{if} ( parent.name contains \{ "day", "month", "year"\} ) \textbf{then} 
      \end{varwidth}
    \State \begin{varwidth}[t]{\linewidth}
      \hskip\algorithmicindent \hskip\algorithmicindent  \hskip\algorithmicindent parentOfParent~$\gets$~parent.getParent()
      \end{varwidth}
    \State \begin{varwidth}[t]{\linewidth}
    \hskip\algorithmicindent \hskip\algorithmicindent \hskip\algorithmicindent \textbf{if} ( parentOfParent is not null \textbf{and} \par 
    \hskip\algorithmicindent \hskip\algorithmicindent \hskip\algorithmicindent \hskip\algorithmicindent parentOfParent.getPosTag==Verb \textbf{and} \par
    \hskip\algorithmicindent \hskip\algorithmicindent \hskip\algorithmicindent \hskip\algorithmicindent parentOfParent.name \textbf{in} investmentVerbsSet )  \textbf{then}
      \end{varwidth}
      \State \begin{varwidth}[t]{\linewidth}
        \hskip\algorithmicindent \hskip\algorithmicindent \hskip\algorithmicindent  \hskip\algorithmicindent  \textbf{return} numberNode
      \end{varwidth}
  \end{algorithmic}
\end{algorithm}

Therefore an utterance like \texttt{"I would like to invest 10 thousands in 3 years"} becomes \texttt{"I would like to invest \#v in \#dt years"}. Or \texttt{"10 in 3 years"} becomes \texttt{"\#v in \#dt years"}, and both intents have the same intent class.

For doing that we implemented a few rules using a depth-first search algorithm combined with the rules as described in Algorithm \ref{getperiod}, Algorithm \ref{getvalue} and Algorithm \ref{parsing}. Note that our parser works only for short texts on which the user's utterance mentions only one period of time and/ or initial amount of investment in the same utterance.

\begin{algorithm}
  \caption{extract\_initial\_amount\_of\_investment (utteranceTree)}\label{getvalue}
  \begin{algorithmic}[1]
    \State \begin{varwidth}[t]{\linewidth}
    numbersNodes~$\gets$~utteranceTree.getNumbersNodes();
      \end{varwidth}
    \State \begin{varwidth}[t]{\linewidth}
    \textbf{foreach}(numberNode \textbf{in} numbersNodes) \textbf{do}
      \end{varwidth}
    \State \begin{varwidth}[t]{\linewidth}
    \hskip\algorithmicindent parentsOfNumbersNode~$\gets$~numbersNode.getParents()
      \end{varwidth}
    \State \begin{varwidth}[t]{\linewidth}
    \hskip\algorithmicindent \textbf{foreach}(parent \textbf{in} parentsOfNumbersNodes)  \textbf{do}
      \end{varwidth}
     \State \begin{varwidth}[t]{\linewidth}
     \hskip\algorithmicindent \hskip\algorithmicindent \textbf{if} ( parent.name \textbf{does not contain} \{ "day", "month", "year"\} ) \textbf{then} 
      \end{varwidth}
      \State \begin{varwidth}[t]{\linewidth}
        \hskip\algorithmicindent \hskip\algorithmicindent \hskip\algorithmicindent   \textbf{return} numberNode
      \end{varwidth}
  \end{algorithmic}
\end{algorithm}

\begin{algorithm}
  \caption{frame\_parsing(utterance, utteranceTree)}\label{parsing}
  \begin{algorithmic}[1]
    \State \begin{varwidth}[t]{\linewidth}
    period~$\gets$~extract\_period\_of\_investment (utteranceTree)
      \end{varwidth}
    \State \begin{varwidth}[t]{\linewidth}
    save\_period\_of\_investment(period)
      \end{varwidth}
    \State \begin{varwidth}[t]{\linewidth}
    value~$\gets$~extract\_initial\_amount\_of\_investment (utteranceTree)
      \end{varwidth}
    \State \begin{varwidth}[t]{\linewidth}
    save\_initial\_amount\_of\_investment(value)
      \end{varwidth}
     \State \begin{varwidth}[t]{\linewidth}
    new\_intent~$\gets$~replace(new\_intent, period, "\#dt")
      \end{varwidth}
    \State \begin{varwidth}[t]{\linewidth}
     new\_intent~$\gets$~replace(new\_intent, value, "\#v")
      \end{varwidth}
  \end{algorithmic}
\end{algorithm}

\subsubsection{Speech Act Classifier}

In CognIA we have complemented the speech act classes with the ones related to the execution of specific actions. Therefore, if the chatbot needed to compute the return of investment, then, once it is computed, the speech act of the reply will be \texttt{Inform\_Calculation} and the one that represents the query for that is \texttt{Query\_Calculation}. In table \ref{tab:cogniaspeechacts} we list the specific ones.
\begin{table}[h]%
\caption{CognIA Specific Speech Acts Classes}
\centering
\begin{tabular}{|p{4.5cm}|p{8cm}|}
\hline
    \textbf{Speech Act}   &   \textbf{Example}        \\ \hline
QUERY DEFINITION &    \textit{What is CD?}   \\ \hline
INFORM DEFINITION &   \textit{A certificate of deposit (CD) is a time deposit, you lend money to the bank so it can use your money to invest somewhere else. }   \\ \hline
QUERY NEWS &    \textit{Nowadays, is is better to invest in CDB or in the Savings Account?}   \\ \hline
INFORM NEWS &   \textit{Depends, but nowadays CDs are good options.}   \\ \hline
QUERY CALCULATION &    \textit{What if I invest 10 thousands for 3 years?}   \\ \hline
INFORM CALCULATION &   \textit{At the end, you will have 12 thousands in the Savings Account and 12,5 thousand with a CD.}   \\ \hline
\end{tabular} 
\vspace{-1\baselineskip}
\label{tab:cogniaspeechacts}
\end{table}%

\subsubsection{Intention Classifier}

Given that there is no public dataset available with financial intents in Portuguese, we have employed the incremental approach to create our own training set for the Intent Classifier. First, we applied the Wizard of Oz method and from this study, we have collected a set of 124 questions that the users asked. Next, after these questions have been manually classified into a set of intent classes, and used to train the first version of the system, this set has been increased both in terms of number of classes and samples per class, resulting in a training set with 37 classes of intents, and a total 415 samples, with samples per class ranging from 3
to 37.

We have defined our classification method based on features extracted from word vectors. Word vectors consist of a way to encode the semantic meaning of the words, based on their frequency of co-occurrence.
To create domain-specific word vectors, a set of thousand documents are needed related to desired domain.
Then each intent from the training set needs to be encoded with its corresponding mean word vector. The mean word vector is then used as feature vector for standard classifiers.

We have created domain-specific word vectors by considering a set 246,945 documents,
corresponding to of 184,001 Twitter posts and and 62,949 news articles, all related to finance
\footnote{We have also evaluated word vectors created with all texts from the Wikipedia in Portuguese,
but after conducting some preliminary experiments we observed that the use of tweets and news articles
presented better results. We believe this happens due to (i) the more specific domain knowledge in the news articles and (ii) the more informal language used in tweets, which tend to be similar to 
texts sent to chatbots.}. 

The set of tweets has been crawled from the feeds of blog users who are considered experts in the 
finance domain. The news article have been extracted from links included in these tweets. 
This set contained a total of 63,270,124 word occurrences, with a vocabulary of 97,616 distinct
words. With the aforementioned word vectors, each intent from the training set has been encoded with
its corresponding mean word vector. The mean word vector has been then used as feature vector
for standard classifiers.

As the base classifier, we have pursued with a two-step approach. In the first step, the main goal
was to make use of a classifier that could be easily retrained to include new classes and intents.
For this reason, the first implementation of the system considered an 1-nearest-neighbor (1NN) 
classifier, which is simply a K-nearest-neighbor classifier with K set to 1. With 1NN, the
developer of the system could simply add new intents and classes to the classifier, by means of
inserting new lines into the database storing the training set. Once we have considered that
the training set was stable enough for the system, we moved the focus to an approach that would
be able to provide higher accuracy rates than 1NN. For this, we have employed Support Vector
Machines (SVM) with a Gaussian kernel, the parameters of which are optimized by means of a
grid search. 

\subsubsection{Action Classifier}

We manually mapped the intent classes used to train the intent classifier to action classes and the dependent entities and features, when the case.
Table \ref{tab:actionclass1} summarizes the number of intent classes per action class that we used in CognIA.

\begin{table}[h]%
\caption{Action Classes Mapping}
\centering
 \begin{tabular}{|p{2.2cm}|p{1.5cm}|p{2.5cm}|p{2.5cm}|}
\hline
   \textbf{Action Class} & \textbf{\#Intent Classes} & \textbf{Entities} & \textbf{Features }   \\ \hline
   Greet                  & 1 &  &  \\ \hline
   Thank 		     & 1 &  &   \\ \hline
   Bye       		     & 1 &  &   \\ \hline
   Get Definition     & 60 &  &   \\ \hline
   Search News     & 45 &  &   \\ \hline
   Compute            & 10 & savings account, certificate of deposit & period of time, initial value \\ \hline
   Ask More            & 79 & savings account, certificate of deposit & period of time, initial value \\ \hline
   Send Information  & 2 &  &  \\ \hline
   Send Refuse     & 1 &  &   \\ \hline
   No Action     & 1 &  &  \\ \hline
\end{tabular} 
\label{tab:actionclass1}
\vspace{-1\baselineskip}
\end{table}%

\subsubsection{Action Execution} 

For the majority of action classes we used SABIA's default behavior. For instance, \texttt{Greet} and \texttt{Bye} actions classes are implemented using rapport, which means that if the user says \textit{"Hi"} the chatbot will reply \textit{"Hi"}. 

The \texttt{Search News}, \texttt{Compute} and \texttt{Ask More} classes are the ones that require specific implemention for CognIA as following:

\begin{itemize}
\item Search News: search finance on social media service \cite{Flavio2016}, \cite{Cavalin2016}  receives the utterance as input, searches on previously indexed Twitter data for finance for Portuguese and return to the one which has the highest score, if found.
\item Ask More: If the user sends an utterance that has the intention class of simulating the return of investment, while not all variables to compute the return of investment are extracted from the dialogue, the mediator keeps asking the user these information before it actually redirects the query to the experts. This action then checks the state of the context given the specified intent flow as described in (\ref{eq:graph}) and (\ref{eq:actionclasseq}) in section \ref{sec:arch} to decide which variables are missing. For CognIA we manually added these dependencies on the database. 

\item Compute: Each expert Chatbot implements this action according to its expertise. The savings account chatbot computes the formula (\ref{form:sa}) and the certificate of deposit computes the formula (\ref{form:cd}). Both are currently formulas for estimating in Brazil.
\begin{equation}
\label{form:sa}
RoI_{SA} = IV + IV*(R+TR)
\end{equation}
where $RoI_{SA}$ is the return of investment for the savings account, $IV$ is the initial value of investment, $R$ is the savings account interest rate and $TR$ is the savings account rate base\footnote{The base rate is the interest rate that is paid each month regardless of whether your investment meets the conditions for the payment of the extra bonus rate.}.
\begin{equation}
\label{form:cd}
RoI_{CD}=IV+(IV*ID*P^d)-IT
\end{equation}
where $RoI_{CD}$ is the return of investment for certificate of deposit, $IV$ is the initial value of investment, $ID$ is the Interbank Deposit rate (DI in Portuguese), $P$ is the ID's percentual payed by the bank (varies from 90\% to 120\%), $d$ is the number of days the money is invested, and finally $IT$ is the income tax on the earnings.
\end{itemize}

\begin{center}
\begin{table}[h]%
\caption{Action Execution} 
   \centering
   \begin{tabular}{|c|p{10cm}|}
   \hline
   \textbf{Action Class} & \textbf{Task and Answer Generation}   \\
   \hline
   \hline
   Greet                  & rapport \\ \hline
   Thank 		     & send \textit{you're welcome}  \\ \hline
   Bye       		     & rapport \\ \hline
   Get Definition     & answer manually pre-defined or search finance on social media server  \\ \hline
   Search News     & search finance on social media server   \\ \hline
   Compute            & loads period of time and initial value from context to compute per chatbot expert \\ \hline
   Ask More            & search on the intent flow graph \\ \hline
   Send Refuse     & answer manually pre-defined  \\ \hline
   No Action     & no answer \\ \hline
   \end{tabular}
 \vspace{-1\baselineskip}
  \label{tab:actionclassexec}
\end{table}%
\end{center}

\subsection{Intention Classifier Accuracy}

In Table~\ref{tab:eval_first_set} we present the comparison of some distinct classification on 
the first version of the training set, i.e. the set used to deploy the first classifier into
the system. Roughly speaking, the 1NN classifier has been able to achieve a level of accuracy
that is higher than other well-known classifiers, such as Logistic Regression and Na\"ive Bayes,
showing that 1NN is suitable as a development classifier. Nevertheless, a SVM can perform
considerable better than 1NN, reaching accuracies of about 12 percentage points higher, which
demonstrates that this type of base classifier is a better choice to be deployed once the 
system is stable enough. It is worth mentioning that these results consider the leave-one-out
validation procedure, given the very low number of samples in some classes.

\begin{table}[h]%
\caption{Evaluation of different classifiers in the first version of the training set}
   \centering
   \begin{tabular}{|c|cccc|}
   \hline
   & {\bf Precision} & {\bf Recall} & {\bf F1} & {\bf Accuracy} \\
   \hline
   \hline
   1NN                 & 0.85 & 0.84 & 0.84 & 0.84 \\
   Logistic Regression & 0.66 & 0.68 & 0.64 & 0.68 \\
   Na\"ive Bayes       & 0.80 & 0.79 & 0.78 & 0.79 \\
   SVM                 & 0.97 & 0.96 & 0.96 & 0.96 \\
   \hline
   \end{tabular}
\vspace{-1\baselineskip}
\label{tab:eval_first_set}
\end{table}%
\begin{table}[h]%
\caption{Evaluation of different classifiers in the most recent version of the training set}
   \centering
   \begin{tabular}{|c|cccc|}
   \hline
   & {\bf Precision} & {\bf Recall} & {\bf F1} & {\bf Accuracy} \\
   \hline
   \hline
   1NN                 & 0.81 & 0.80 & 0.80 & 0.80 \\
   Logistic Regression & 0.75 & 0.77 & 0.75 & 0.77 \\
   Na\"ive Bayes       & 0.79 & 0.77 & 0.76 & 0.77 \\
   SVM                 & 0.96 & 0.95 & 0.95 & 0.95 \\
   \hline
   \end{tabular}
\vspace{-1\baselineskip}
\label{tab:eval_most_recent_set}
\end{table}%

As we mentioned, the use of an 1NN classifier has allowed the developer of the system to easily add
new intent classes and samples whenever they judged it necessary, so that the system could present
new actions, or the understanding of the intents could be improved. As a consequence, the initial
training set grew from 37 to 63 classes, and from 415 to 659 samples, with the number of samples
per class varying from 2 to 63. For visualizing the impact on the accuracy of the system, in 
Table~\ref{tab:eval_most_recent_set} we present the accuracy of the same classifiers used in the
previous evaluation, in the new set. In this case, we observe some drop in accuracy for 1NN,
showing that this classifier suffers in dealing with scalability. On the other hand, SVM has shown
to scale very well to more classes and samples, since its accuracy kept at a very similar level
than that with the other set, with a difference of about only 1 percentage point.

\subsection{Testing SABIA}

In this section, we describe the validation framework that we created for integration tests. For this, we developed it as a new component of SABIA's system architecture and it provides a high level language which is able to specify interaction scenarios that simulate users interacting with the deployed chatbots. The system testers provide a set of utterances and their corresponding expected responses, and the framework automatically simulates users interacting with the bots and collect metrics, such as time taken to answer an utterance and other resource consumption metrics (e.g., memory, CPU, network bandwidth). Our goal was to: (i) provide a tool for integration tests, (ii) to validate CognIA's implementation, and (iii) to support the system developers in understanding the behavior of the system and which aspects can be improved. Thus, whenever developers modify the system's source code, the modifications must first pass the automatic test before actual deployment.

\subsubsection{Test framework}

The test framework works as follows. The system testers provide a set $D = \{d_1, ..., d_n\}$ of dialogues as input. Each dialogue $d_i$ $\in$ $D$ is an ordered set whose elements are represented by $<u, R>$, where $u$ is the user utterance and $R$ is an ordered set of pairs $<r, c>$ that lists each response $r$ each chatbot $c$ should respond when the user says $u$. For instance, Table~\ref{tab:dialog_test_1} shows a typical dialogue ($d_1$) between a user and the CognIA system. Note that we are omitting part of the expected answer with "..." just to better visualize the content of the table.

\begin{center}
\begin{longtable}{|p{3.6cm}p{0.4cm}p{4.5cm}p{3.2cm}|}%
\caption[Content of dialogue $d_1$ (example of dialogue in CognIA]{Content of dialogue $d_1$ (example of dialogue in CognIA)} \label{tab:dialog_test_1} \\
   \hline
 {\bf User utterance $u$} & {\bf rId} & {\bf Expected response $r$} & {\bf Chatbot $c$} \\
   \hline
   \hline

\rowcolor{gray!25}	 hello							&1& Hello 											& Mediator  			\\ \hline
\rowcolor{white}	 what is cdb?						&2& @CDBExpert	 what is cdb?							& Mediator			\\
\rowcolor{white}									&3& CDB is a type of investment that...				 		& CDB Expert			\\ \hline
 \rowcolor{gray!25} 	which is better: cdb or savings account? 	&4& I found a post in the social media for....					& Mediator  			\\ \hline
 \rowcolor{white} 	i would like to invest R\$ 50 in six months 	&5& @SavingsAccountExpert and @CDBExpert, could you do a simulation...	& Mediator  			\\
 \rowcolor{white} 								 	&6& If you invest in Savings Account, ...						& Savings Account Exp. 	\\
 \rowcolor{white} 								 	&7& If you invest in CDB,... 								& CDB Expert 			\\
 \rowcolor{white} 								 	&8& Thanks											& Mediator 			\\
 \rowcolor{white} 								 	&9& @User, there is no significant difference..					& Mediator 			\\ \hline
 \rowcolor{gray!25} 	so i want to invest R\$ 10000 in 2 years	&10& @SavingsAccountExpert and @CDBExpert, could you do a simulation...			& Mediator  			\\
 \rowcolor{gray!25} 								 	&11& If you invest in Savings Account,... 						& Savings Account Exp. 	\\
 \rowcolor{gray!25} 								 	&12& If you invest in CDB,... 								& CDB Expert 			\\
 \rowcolor{gray!25} 								 	&13& Thanks											& Mediator 			\\
 \rowcolor{gray!25} 								 	&14& @User, in that case, it is better...						& Mediator 			\\ \hline  
 \rowcolor{white} 	what if i invest R\$10,000 in 5 years? 	&15& @SavingsAccountExpert and @CDBExpert, could you do a simulation...			& Mediator  			\\
 \rowcolor{white} 									&16& If you invest in Saving Account,... 						& Savings Account Exp. 	\\
 \rowcolor{white} 								 	&17& If you invest in CDB,... 								& CDB Expert 			\\
 \rowcolor{white} 								 	&18& Thanks											& Mediator 			\\
 \rowcolor{white} 								 	&19& @User, in that case, it is better...						& Mediator 			\\ \hline
 \rowcolor{gray!25} 	how about 15 years?					&20& @SavingsAccountExpert and @CDBExpert, could you do a simulation...				& Mediator  			\\
 \rowcolor{gray!25} 								 	&21& If you invest in Savings Account,... 						& Savings Account Exp 	\\
 \rowcolor{gray!25} 								 	&22& If you invest in CDB,... 								& CDB Expert 			\\
 \rowcolor{gray!25} 								 	&23& Thanks											& Mediator 			\\
 \rowcolor{gray!25} 								 	&24& @User, in that case, it is better...						& Mediator 			\\ \hline
 \rowcolor{white} 	and 50,0000?						&25& @SavingsAccountExpert and @CDBExpert, could you do a simulation...				& Mediator  			\\
 \rowcolor{white} 								 	&26& If you invest in Savings Account,... 						& Savings Account Exp. 	\\
 \rowcolor{white} 								 	&27& If you invest in CDB,... 								& CDB Expert 			\\
 \rowcolor{white} 								 	&28& Thanks											& Mediator 			\\
 \rowcolor{white} 								 	&29& @User, in that case, it is better..						& Mediator 			\\ \hline
 \rowcolor{gray!25} 	I want to invest in 50,000 for 15 years in CDB	&30& Sure, follow this link to your bank...					& Mediator  			\\ \hline
 \rowcolor{white} 	thanks								&31& You are welcome.								& Mediator 			\\ \hline
\end{longtable}%
\end{center}

The testers may also inform the number of simulated users that will concurrently use the platform. Then, for each simulated user, the test framework iterates over the dialogues in $D$ and iterates over the elements in each dialogue to check if each utterance $u$ was correctly responded with $r$ by the chatbot $c$. There is a maximum time to wait. If a bot does not respond with the expected response in the maximum time (defined by the system developers), an error is raised and the test is stopped to inform the developers about the error. Otherwise, for each correct bot response, the test framework collects the time taken to respond that specific utterance by the bot for that specific user and continues for the next user utterance. Other consumption resource metrics (memory, CPU, network, disk). The framework is divided into two parts. One part is responsible to gather resource consumption metrics and it resides inside SABIA. The other part works as clients (users) interacting with the server. It collects information about time taken to answer utterances and checks if the utterances are answered correctly.

By doing this, we not only provide a sanity test for the domain application (CognIA) developed in SABIA framework, but also a performance analysis of the platform. That is, we can: validate if the bots are answering correctly given a pre-defined set of known dialogues, check if they are answering in a reasonable time, and verify the amount of computing resources that were consumed to answer a specific utterance. Given the complexity of CognIA, these tests enable debugging of specific features like: understanding the amount of network bandwidth to use external services, or analyzing CPU and memory consumption when responding a specific utterance. The later may happen when the system is performing more complex calculations to indicate the investment return, for instance.

\subsubsection{Test setup}

CognIA was deployed on IBM Bluemix, a platform as a service, on a Liberty for Java Cloud Foundry app with 3 GB RAM memory and 1 GB disk. Each of the modules shown in Figure \ref{fig:cognia} are deployed on separate Bluemix servers. \texttt{Node.JS} and \texttt{Socket.IO} servers are both deployed as Node Cloud Foundry apps, with 256 MB RAM memory and 512 MB disk each. \texttt{Search Finance on Social Media} is on a Go build pack Cloud Foundry app with 128 MB RAM memory and 128 GB disk. For the framework part that simulates clients, we instantiated a virtual machine with 8 cores on IBM's SoftLayer that is able to communicate with Bluemix. Then, the system testers built two dialogues, i.e., $D = \{d_1, d_2\}$. The example shown in Table~\ref{tab:dialog_test_1} is the dialogue test $d_1$. For the dialogue $d_2$, although it also has 10 utterances, the testers varied some of them to check if other utterances in the finance domain (different from the ones in dialogue $d_1$) are being responded as expected by the bots. Then, two tests are performed and the results are analyzed next. All tests were repeated until the standard deviation of the values was less than 1\%. The results presented next are the average of these values within the 1\% margin.

\subsubsection{Results}

 \textbf{Test 1}: The first test consists of running both dialogues $d_1$ and $d_2$ for only one user for sanity check. We set \texttt{30 seconds} as the maximum time a simulated user should wait for a bot correct response before raising an error. The result is that all chatbots (\texttt{Mediator, CDBExpert, and SavingsAccountExpert}) responded all expected responses before the maximum time. Additionally, the framework collected how long each chatbot took to respond an expected answer. 
 
\begin{figure}[h]
\centerline{\includegraphics[width=\textwidth]{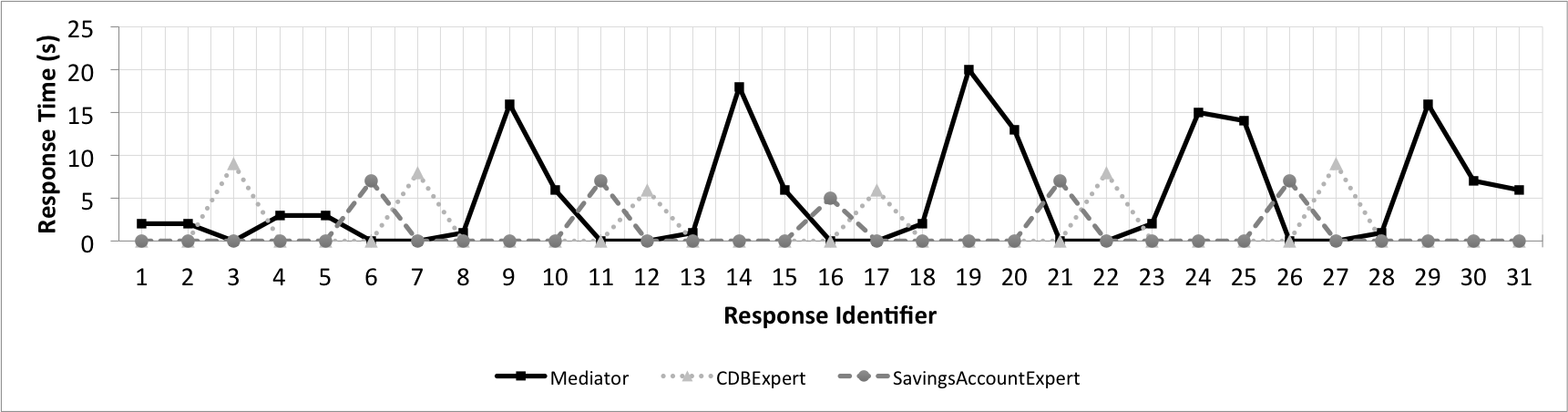}}
\caption{Single simulated user results for dialogue $d_1$.}
\label{fig:single-user}
\end{figure}

In Figure \ref{fig:single-user}, we show the results for those time measurements for dialogue $d_1$, as for the dialogue $d_2$ the results are approximately the same. The x-axis (\texttt{Response Identifier}) corresponds to the second column (\texttt{Resp. Id}) in Table~\ref{tab:dialog_test_1}. We can see, for example, that when the bot \texttt{CDBExpert} responds with the message \texttt{3} to the user utterance \texttt{"what is cdb?"}, it is the only bot that takes time different than zero to answer, which is the expected behavior. We can also see that the \texttt{Mediator} bot is the one that takes the longest, as it is responsible to coordinate the other bots and the entire dialogue with the user. Moreover, when the expert bots (\texttt{CDBExpert and SavingsAccountExpert}) are called by the \texttt{Mediator} to respond to the simulation calculations (this happens in responses \texttt{6, 7,  11, 12,  16, 17,  21, 22,  26, 27}), they take approximately the same to respond. Finally, we see that when the concluding responses to the simulation calculations are given by the \texttt{Mediator} (this happens in responses \texttt{9, 14, 19, 24, 29}), the response times reaches the greatest values, being 20 seconds the greatest value in response \texttt{19}. These results support the system developers to understand the behavior of the system when simulated users interact with it and then focus on specific messages that are taking longer.

\begin{figure}[h]
\centerline{\includegraphics[width=\textwidth]{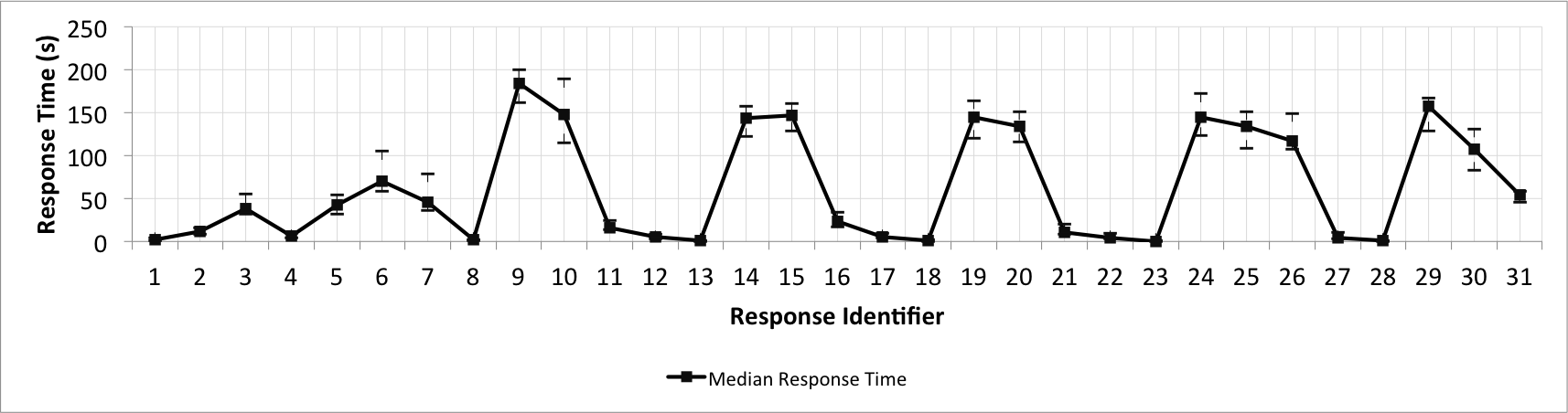}}
\caption{Eight concurrent simulated users.}
\label{fig:eight-users}
\end{figure}
 
 \textbf{Test 2}: This test consists of running dialogue $d_1$, but now using eight concurrent simulated users. We set the maximum time to wait to \texttt{240 seconds}, i.e., eight times the maximum set up for the single user in Test 1. The results are illustrated in Figure \ref{fig:eight-users}, where we show the median time for the eight users. The maximum and minimum values are also presented with horizontal markers. Note that differently than what has been shown in Figure \ref{fig:single-user}, where each series represents one specific chatbot, in Figure \ref{fig:eight-users}, the series represents the median response time for the responses in the order (x-axis) they are responded, regardless the chatbot. 

Comparing the results in Figure \ref{fig:eight-users} with the ones in Figure \ref{fig:single-user}, we can see that the bots take longer to respond when eight users are concurrently using the platform than when a single user uses it, as expected. For example, \texttt{CDBExpert} takes approximately 5 times longer to respond response \texttt{3} to eight users than to respond to one user. On average, the concluding responses to the simulation questions (i.e., responses \texttt{9, 14, 19, 24, 29}) take approximately 7.3 times more to be responded with eight users than with one user, being the response \texttt{9} the one that presented greatest difference (11.4 times longer with eight users than with one). These results help the system developers to diagnose the scalability of the system architecture and to plan sizing and improvements.

\section{Conclusions and Future Work}

In this article, we explored the challenges of engineering \gls{MPCS} and we have presented a hybrid conceptual architecture and its implementation with a finance advisory system. 

We are currently evolving this architecture to be able to support decoupled interaction norms specification, and we are also developing a multi-party governance service that uses that specification to enforce exchange of compliant utterances. 

In addition, we are exploring a micro-service implementation of SABIA in order to increase its scalability and performance, so thousands of members can join the system within thousands of conversations.

\appendix
\section*{Acknowledgments}
The authors would like to thank Maximilien de Bayser, Ana Paula Appel, Flavio Figueiredo and Marisa Vasconcellos, who contributed with discussions during SABIA and CognIA's implementation.


\label{lastpage}
\end{document}